\documentclass[12pt,a4paper, notitlepage]{article}
\usepackage[english]{babel}
\usepackage[utf8]{inputenc}
\usepackage{sectsty}
\sectionfont{\color{blue}}
\subsectionfont{\color{red}}
\usepackage{etoolbox}
\usepackage[colorinlistoftodos]{todonotes}
\usepackage{graphicx}
\usepackage{caption}
\usepackage{amsmath}
\usepackage{mathtools}
\usepackage{pifont}
\usepackage[toc]{appendix}
\usepackage{multirow}
\usepackage{amsfonts}
\usepackage[normalem]{ulem}
\usepackage{amsmath}
\usepackage{multirow}
\usepackage{float}
\usepackage[table]{xcolor}
\usepackage[framemethod=TikZ]{mdframed}
\mdfdefinestyle{MyFrame}{%
	linecolor=blue,
	outerlinewidth=2pt,
	roundcorner=20pt,
	innertopmargin=\baselineskip,
	innerbottommargin=\baselineskip,
	innerrightmargin=20pt,
	innerleftmargin=20pt,
	backgroundcolor=gray!50!white}
\newcommand\xoutpars[1]{\let\helpcmd\xout\parhelp#1\par\relax\relax}
\newcommand\soutpars[1]{\let\helpcmd\sout\parhelp#1\par\relax\relax}
\long\def\parhelp#1\par#2\relax{%
	\helpcmd{#1}\ifx\relax#2\else\par\parhelp#2\relax\fi%
}

\usepackage{tabularx,booktabs,ragged2e,caption}
\usepackage{lmodern}
\usepackage{lineno}
\usepackage{amsmath,amssymb,amsfonts,systeme}
\usepackage{makecell}
\usepackage[T1]{fontenc}
\usepackage{lmodern}
\usepackage{amsmath,amssymb,systeme}
\usepackage{makecell}
\usepackage{soul}
\usepackage{subcaption}
\usepackage{threeparttable}
\usepackage{booktabs}
\usepackage{url}
\usepackage[round]{natbib}
\usepackage{listings}
\usepackage{xcolor}
\usepackage{array}
\usepackage{caption}
\usepackage{CJKutf8}
\definecolor{mauve}{rgb}{0.58,0,0.82}
\definecolor{dkgreen}{rgb}{0,0.45,0}
\usepackage{url}

\usepackage{siunitx}
\usepackage{xcolor}
\usepackage{booktabs,colortbl, array}
\usepackage{pgfplots}
\usepackage{pgfplotstable}
\pgfplotsset{compat=1.8}
\newcommand{\safeincludegraphics}[2][]{%
	\IfFileExists{#2}{%
		\includegraphics[#1]{#2}%
	}{%
		\fbox{\parbox[c][0.25\textheight][c]{0.8\linewidth}{\centering Missing figure file:\\ \texttt{\detokenize{#2}}}}%
	}%
}

\definecolor{rulecolor}{RGB}{0,71,171}
\definecolor{tableheadcolor}{gray}{0.92}
\newcommand{\topline}{ %
	\arrayrulecolor{rulecolor}\specialrule{0.1em}{\abovetopsep}{0pt}%
	\arrayrulecolor{tableheadcolor}\specialrule{\belowrulesep}{0pt}{0pt}%
	\arrayrulecolor{rulecolor}}
\newcommand{\midtopline}{ %
	\arrayrulecolor{tableheadcolor}\specialrule{\aboverulesep}{0pt}{0pt}%
	\arrayrulecolor{rulecolor}\specialrule{\lightrulewidth}{0pt}{0pt}%
	\arrayrulecolor{white}\specialrule{\belowrulesep}{0pt}{0pt}%
	\arrayrulecolor{rulecolor}}
\newcommand{\bottomline}{ %
	\arrayrulecolor{white}\specialrule{\aboverulesep}{0pt}{0pt}%
	\arrayrulecolor{rulecolor} %
	\specialrule{\heavyrulewidth}{0pt}{\belowbottomsep}}%

\pgfplotstableset{normal/.style ={%
		header=true,
		string type,
		column type=l,
		every odd row/.style={
			before row=
		},
		every head row/.style={
			before row={\topline\rowcolor{tableheadcolor}},
			after row={\midtopline}
		},
		every last row/.style={
			after row=\bottomline
		},
		col sep=&,
		row sep=\\
	}
}
\usepackage[]{mdframed}
\usepackage{tikz}
\usetikzlibrary{fit}

\usepackage{blkarray}
\usepackage{enumitem,lipsum}
\usepackage[round]{natbib}
\usepackage{setspace}
\newenvironment{keywords}{\par\noindent\small}{\par}

\usepackage{hyperref}
\hypersetup{colorlinks,linkcolor={blue},citecolor={blue},urlcolor={red}}

\usepackage{amssymb}
\usepackage{pifont}%
\usepackage{makecell}
\usepackage{lineno}
\title{Contextual Semantic Relevance and Word Surprisal Predict N400 and P600 Dynamics During Naturalistic Reading}

\date{\vspace{-5ex}} 

\author{
	Kun Sun \\
	\small Tongji University \\
	\small \texttt{kunsun@tongji.edu.cn}
	\and
	Rong Wang \\
	\small University of Tuebingen \\
	\small \texttt{rong.wang@uni-tuebingen.de}
}

\begin{document}

\maketitle

\begin{abstract}

Word surprisal is a well-established computational predictor of human neural responses during language comprehension, but it remains less clear whether local semantic fit explains neural response variation beyond lexical expectation during naturalistic reading.
Using the Dublin EEG-based Reading Experiment Corpus (DERCo), this
study examined whether contextual semantic relevance predicts word-locked EEG activity in the N400 and P600 windows. Contextual semantic relevance was computed as an attention-aware measure of how strongly a target word is semantically connected to its recent discourse context, and it was compared with GPT-based word surprisal. Across 22 participants and 32 EEG channels, we tested both predictors using regression-based ERP analyses and generalized additive mixed models while controlling for lexical variables and repeated observations.
Both predictors were reliably associated with EEG responses, but they showed partly different temporal and scalp-level patterns. Surprisal captured expectancy-related variation, whereas contextual semantic relevance showed robust effects across N400- and P600-window mean voltages, with particularly strong explanatory support in the P600 window. Model comparisons indicated that contextual semantic relevance contributed explanatory value beyond lexical controls and surprisal. These findings suggest that naturalistic reading depends
on both lexical expectation and local semantic integration, and that contextual semantic relevance offers an interpretable computational link between discourse semantic fit and ERP dynamics.
  	  
\end{abstract}


\newpage

\section{Introduction}

Computational models provide a powerful way to connect linguistic information with human neural activity. In language research, word surprisal has become a central expectation based metric because it quantifies how unexpected a word is given its preceding context (\citealp{hale2001probabilistic}; \citealp{levy2008expectation}). In contrast, memory-based accounts emphasize the storage, retrieval, and integration of prior input, as in ACT-R and short term memory models (\citealp{anderson1997act}; \citealp{ritter2019act}; \citep{baddeley2010working}). Within this broader memory based perspective, semantic similarity and semantic relevance provide measures of the semantic relation between words and contexts \citep{mitchell2010composition,hollis2016principals}.

Word surprisal has strong empirical support. Less predictable words are generally more difficult to process, as shown in reading times, self paced reading, eye tracking, and neuroimaging studies (\citealp{demberg2008data}; \citealp{smith2013effect}; \citealp{hale2016information}; \citealp{shain2020fmri}; \citealp{rayner1998eye}; \citealp{boston2008parsing}). Memory-based theories offer a complementary view in which processing difficulty arises from retrieving and integrating context with new input (\citealp{gibson1998linguistic}). Contextual semantic relevance extends this idea by estimating how strongly a target word fits its surrounding discourse context. This metric has shown promise in modeling word processing during naturalistic comprehension (\citealp{roland2012semantic}; \citealp{sun2023interpretable}; \citealp{sun2024attention}).

Both surprisal and semantic relevance have also been linked to neural activity during language comprehension. Surprisal predicts the ERP component \citep{frank2015erp} and has been associated with EMG and fMRI responses (\citealp{brennan2016naturalistic}; \citealp{willems2015prediction}; \citealp{hale2022neurocomputational}; \citealp{goldstein2022shared}). Semantic metrics have likewise been used to explain neural responses to language (\citealp{frank2017word}; \citealp{broderick2019semantic}). These findings suggest that lexical expectation and semantic integration may both contribute to the neural dynamics of comprehension.

The N400 and P600 provide important temporal markers of language related neural processing. The N400 is a negative response around 400 ms after word presentation \citet{kutas1980reading}, whereas the P600 is a later positive response around 600 ms \citet{hagoort1999neurocognition}. The N400 has been linked both to predictive preactivation (\citealp{kutas2011thirty}; \citealp{szewczyk2018n400}) and to context sensitive retrieval or semantic integration (\citealp{lau2013dissociating}; \citealp{federmeier2022connecting}; \citealp{delogu2021components}). Similar issues arise for the P600, which has been associated with prediction violations (\citealp{kuperberg2007neural}; \citealp{qi2017native}) as well as semantic integration in the semantic P600 literature (\citealp{van2006heuristics}; \citealp{kuperberg2007neural}). These findings raise a central question for language neuroscience: How do lexical expectation and semantic integration jointly shape the N400 and P600 during naturalistic comprehension?

This question is also relevant to theories of processing architecture. Sequential accounts propose that lexical semantic access indexed by the N400 precedes later integration indexed by the P600. Parallel accounts propose that multiple information sources can influence both components (\citealp{brouwer2012getting}; \citealp{brouwer2017neurocomputational}). Related debates also appear in models of reading, including E Z Reader and SWIFT (\citealp{pollatsek2006tests}; \citealp{engbert2005swift}). Recent work has called for more unified accounts of how prediction, retrieval, and integration unfold over time (\citealp{michaelov2024strong}; \citealp{lopopolo2024tracking}).

Despite this progress, several limitations remain. First, surprisal and semantic relevance have often been tested separately or over a small number of EEG channels \citep{frank2015erp}. Second, relatively few studies have examined both metrics for both N400 and P600 responses during naturalistic reading. Third, existing semantic relevance metrics require greater interpretability and validation \citep{sun2023interpretable}. Fourth, much prior work has relied on isolated sentences or listening materials, leaving open how these metrics explain neural activity during naturalistic discourse reading \citep{broderick2019semantic}. Finally, simple correlation analyses or simple regression analyses are insufficient for estimating the unique contributions of multiple predictors to ERP responses, especially when control predictors and participant variability should be modeled in statistical analysis (\citealp{michaelov2024strong}).

The present study addresses these limitations by analyzing word-locked EEG responses during naturalistic reading. We compute word surprisal and proposed ``attention-aware'' method to calculate contextual semantic relevance metric respectively, and we examine their effects on N400 and P600 size across 32 EEG channels. We also include word frequency, word length and participants as control predictors. To test these effects, we combine regression-based ERP analysis with generalized additive mixed models as statistical analysis tools. This design allows us to evaluate how expectation and semantic integration contribute to the temporal and spatial dynamics of language comprehension. 

The study addresses two research questions.

\begin{itemize}
	\item[] 1) How do word surprisal and contextual semantic relevance influence N400 and P600 responses across EEG channels during naturalistic discourse reading?
	\item[] 2) Do these metrics make distinguishable contributions to N400 and P600 dynamics, and what do these contributions suggest about expectation and semantic integration in human language processing?
\end{itemize}

\section{Related Work}

\subsection{Word surprisal and contextual semantic relevance}

\textit{Word surprisal} quantifies how unexpected a word is given its preceding context. It is defined as the negative logarithm of the conditional probability of a word,  formalized as: $S(w_i) = -\log P(w_i|w_1,...,w_{i-1})$, where $S(w_i)$ is the surprisal of word $w_i$ and $P(w_i|w_1,...,w_{i-1})$ is its probability given the preceding words. Surprisal has been widely used to model word processing because less predictable words generally require greater processing effort \citep{hale2016information}. Recent advances in language modeling have improved the estimation of word probabilities, moving from n-gram models to Transformer-based architectures (\citealp{wolf2020transformers}; \citealp{goodkind2018predictive}; \citealp{wilcox2020predictive}; \citealp{schrimpf2021neural}). Autoregressive language models such as \texttt{GPT} provide a natural framework for computing word surprisal because they assign probabilities to upcoming tokens based on preceding context.

Surprisal, as a well-established computational metric, has also been linked to neural responses during language comprehension. It predicts the N400 component and has been associated with EMG and fMRI responses (\citealp{frank2015erp}; \citealp{brennan2016naturalistic}; \citealp{henderson2016language}; \citealp{willems2015prediction}; \citealp{hale2022neurocomputational}; \citealp{goldstein2022shared}). However, it still remains unclear how surprisal influences both N400 and P600 responses across broader EEG channel coverage during naturalistic reading, particularity using advanced statistical analysis. 

Contextual semantic relevance could also provide a critical way to model language comprehension. A large body of work has shown that semantic plausibility facilitates reading and comprehension, but many early studies relied on human ratings of semantic relatedness or plausibility (\citealp{hohenstein2010semantic}; \citealp{delong2014predictability}; \citealp{veldre2016semantic}). Although these ratings are informative, they are subjective and difficult to scale to large naturalistic texts. Computational semantic similarity metrics offer a more scalable alternative. In the present study, semantic similarity refers to the relation between two words or linguistic units, whereas contextual semantic relevance refers to the degree to which a target word is semantically related to its surrounding context. Semantic similarity has been widely used in natural language processing, cognitive psychology, and artificial intelligence (\citealp{jabeen2020semantic}; \citealp{harispe2022semantic}), and it has been shown to predict eye movements and neural responses during language comprehension (\citealp{pereira2018toward}; \citealp{broderick2018electrophysiological}; \citealp{hale2022neurocomputational}).

Several approaches have been proposed to compute contextual semantic relevance. \citet{frank2017word} computed the cosine similarity between the target word vector and the sum of preceding word vectors. \citet{michaelov2024strong} used the cosine similarity between the target word vector and the mean vector of the context, which is closely related to this earlier approach. \citet{broderick2019semantic} used a Euclidean similarity approach, although such metrics may be sensitive to high dimensional vector properties. The limitations of these approaches have been discussed by \citet{sun2023interpretable} and \citet{sun2024attention}. To improve interpretability and contextual sensitivity, \citet{sun2023interpretable} proposed a modified cosine approach that includes both content and function words. More recent ``attention-aware'' approaches incorporate contextual information and distance-sensitive weighting to compute more robust contextual semantic relevance metrics (\citealp{sun2023interpretable}; \citealp{sun2024attention1}). Because this ``attention-aware'' method has shown better performance than earlier cosine and dynamic approaches, the present study adopts it to compute contextual semantic relevance.

\subsection{N400 vs. P600 and processing models}

The functional interpretation of the N400 remains debated. One view, predictive preactivation, proposes that N400 amplitude reflects the conditional probability of a word in context. Expected words require less activation when encountered and therefore elicit smaller N400 responses (\citealp{kutas2011thirty}; \citealp{szewczyk2018n400}). A second view proposes that the N400 reflects context-sensitive retrieval or semantic integration, with amplitude determined by the relation between the incoming word and the semantic features of prior context (\citealp{Lau2008n400}; \citealp{lau2013dissociating}; \citealp{federmeier2022connecting}; \citealp{delogu2021components}). Both views have empirical support, but they often emphasize different subsets of N400 effects. Recent work has therefore called for more integrative accounts (\citealp{michaelov2024strong}; \citealp{lopopolo2024tracking}).

Similar issues arise for the P600. The P600 has traditionally been linked to syntactic prediction, reanalysis, and integration difficulty (\citealp{kuperberg2007neural}; \citealp{qi2017native}). However, the semantic P600 literature shows that semantic anomalies can also elicit P600 responses, suggesting that the P600 is sensitive to semantic coherence as well as syntax (\citealp{van2006heuristics}; \citealp{kuperberg2007neural}). Cloze probability also influences both N400 and P600 responses. Higher cloze probability words elicit smaller N400 responses (\citealp{kutas1984brain}; \citealp{van1990interactions}), whereas lower predictability can increase P600 amplitudes when integration demands are high (\citealp{federmeier1999rose}; \citealp{hagoort2004integration}; \citealp{aurnhammer2023p600}).

These findings raise a broader question about whether language processing proceeds sequentially or in parallel. Sequential or single-stream models propose that lexical retrieval and expectation are reflected in the N400, followed by later discourse integration reflected in the P600 \citep{aurnhammer2021retrieval}. In contrast, parallel or multi-stream models propose that semantic, syntactic, and contextual information can influence processing concurrently, with the N400 and P600 reflecting partially overlapping but distinguishable processes \citep{brouwer2012getting}. This debate is related to broader accounts of cognition and reading, including serial and parallel processing theories (\citealp{townsend1990serial}; \citealp{sigman2008brain}; \citealp{brodbeck2022parallel}) and reading models such as E-Z Reader and SWIFT (\citealp{pollatsek2006tests}; \citealp{engbert2005swift}; \citealp{sun2024attention1}).

Semantic relevance is especially useful for evaluating these accounts because semantic coherence influences both N400 and P600 responses. Semantically incongruent words elicit larger N400 amplitudes, reflecting increased semantic processing demands (\citealp{kutas1980reading}; \citealp{kutas2011thirty}). Semantic anomalies can also elicit P600 responses, reflecting additional processing needed to restore coherence or resolve conflict (\citealp{osterhout1992event}; \citealp{hagoort2003brain}). However, many previous studies examined these effects using isolated sentence materials, or statistical models without sufficient control predictors and random effects. The present study therefore examines surprisal and contextual semantic relevance simultaneously across 32 EEG channels during naturalistic reading, using two statistical models with control predictors and participant-level random effects.

\subsection{Other predictors}

\textit{Word frequency} is a robust predictor of language processing. High-frequency words typically elicit smaller or more positive N400 responses than low-frequency words, reflecting easier lexical retrieval and immediate semantic access (\citealp{van1990interactions}, \citealp{rugg1990event}). Frequency effects are strongest when contextual constraints are weak, but they can be reduced when semantic or syntactic context strongly constrains interpretation. The P600 may also be modulated by word frequency, since low-frequency words can require greater effort to integrate into the ongoing context (\citealp{regel2014distinguishing}; \citealp{qi2017native}).

\textit{Word length} also affects language processing. Longer words can elicit larger N400 responses, likely reflecting greater orthographic, phonological, or lexical processing demands \citep{schuster2016words}. Word length may also modulate the P600 when longer words increase syntactic or semantic integration effort (\citealp{chen2013context}; \citealp{laszlo2011n400}). As word frequency and word length are established predictors of ERP responses and other language comprehension studies, the present study includes both as control predictors in the mixed-effects models.


\section{Methods}

\subsection{Dataset and EEG preprocessing}
\label{sec:data_preprocessing}


We analyzed the EEG subset of the Dublin EEG-based Reading Experiment Corpus (\texttt{DERCo}) \citep{quach2024derco}, a publicly available naturalistic reading dataset. \texttt{DERCo} combines behavioural next-word prediction data from 500 online participants with EEG recordings from healthy adult native English speakers. The 22 EEG participants read five short narrative texts presented word by word in a rapid serial visual presentation (RSVP) paradigm.

Each trial began with a fixation cross. Words were presented in lowercase white letters on a gray background for 200 ms, followed by a 300 ms blank screen. Word onsets were marked using a photodiode-based trigger system, allowing precise word-locked epoching. Thus, the epochs analyzed here were time-locked to word onset rather than to fixation events.

\begin{figure*}[htp]
	\vskip -0.06in
	\centering
	
	\safeincludegraphics[width=0.7\textwidth]{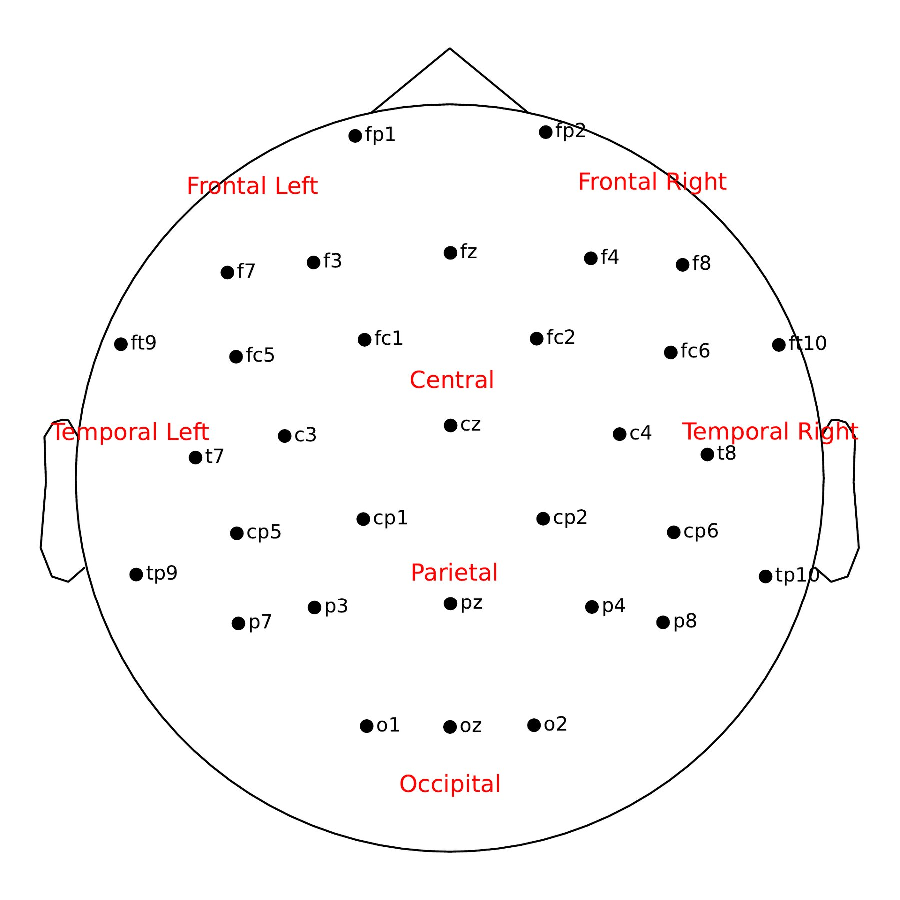}
	
	\caption{The EEG Topomap with regions}
	
	\label{fig:eegmap}
	\vskip -0.11in	
\end{figure*}

EEG was recorded from 32 scalp channels using an ActiCHamp system with an active electrode cap arranged according to the international 10--20 system (see Table~\ref{table:eeg_channel_groups} and Figure~\ref{fig:eegmap}). Signals were sampled at 1,000 Hz and recorded with the amplifier internal reference. Electrode impedances were kept below 5 k$\Omega$. The 32 electrodes were used for channel-wise analyses and were also grouped into scalp regions for regional summaries. As scalp EEG is affected by volume conduction, these channel and regional groupings are interpreted as scalp-level distributions rather than as evidence for specific cortical sources.

The present study used the released preprocessed word-locked epoch files from \texttt{DERCo}. According to the dataset documentation and preprocessing code, the EEG data were filtered with a 0.1--45 Hz bandpass filter, and 50 Hz line-noise filtering was applied. The data were epoched around word-onset events, assigned a standard 10--20 montage, and re-referenced to the common average. Artifact handling
included FASTER-style global outlier detection, ICA with PCA pre-whitening, Picard ICA, ICLabel-based artifact-component removal, and final Autoreject cleaning. Because these steps had already been applied in the released files, no additional filtering, ICA, bad-channel interpolation, or artifact rejection was performed in the present analysis.

The preprocessed epochs were loaded in MNE-Python \citep{gramfort2014mne}. Each participant contributed up to five article-level epoch files, corresponding to the five stories. Word epochs were aligned with token-level predictors by matching the \texttt{DERCo} event labels to the study design matrix. 
The released epochs covered the interval from -200 ms to 1000 ms relative to word onset and contained 1201 samples at 1000 Hz. The epoch construction code specifies \texttt{baseline=None}, and the released \texttt{.fif} files did not store an explicit MNE baseline-correction setting. We therefore analyzed the released cleaned voltages without applying an additional prestimulus baseline correction. This choice is appropriate for the RSVP design because the -200--0 ms interval may contain residual activity from preceding words. Accordingly, the dependent variables are interpreted as window-based mean voltages rather than baseline-corrected ERP amplitudes.

\begin{table}[H]
    \centering
    \caption{EEG channel classifications in the DERCo and their relevance to language and speech processing. The functions listed are literature-based associations reported for these scalp locations; because of volume conduction they do not imply that the signal at a channel originates from the cited cortical region}
    \label{table:eeg_channel_groups}

    \small
    \setlength{\tabcolsep}{5pt}
    \renewcommand{\arraystretch}{1.15}

    \begin{tabularx}{\textwidth}{p{3.0cm} p{4.8cm} X}
        \toprule
        \textbf{Scalp group} & \textbf{Channels} & \textbf{Use in the present analysis} \\
        \midrule

        Frontal &
        Fp1, Fp2, F3, F4, F7, F8, Fz &
        Used to describe anterior scalp activity and possible frontal contributions to language-related EEG responses. \\

        Fronto-central / fronto-temporal &
        FC1, FC2, FC5, FC6, FT9, FT10 &
        Used to summarize activity over fronto-central and lateral scalp sites. \\

        Central / temporal &
        C3, C4, Cz, T7, T8 &
        Used to describe central and lateral scalp activity during word-level reading responses. \\

        Centro-parietal / parietal &
        CP1, CP2, CP5, CP6, P3, P4, P7, P8, Pz &
        Used to summarize posterior scalp activity, including the centro-parietal distribution commonly associated with N400 effects. \\

        Occipital / posterior temporal &
        O1, O2, Oz, TP9, TP10 &
        Used to describe posterior scalp activity related to visual word presentation and later posterior EEG responses. \\

        \bottomrule
    \end{tabularx}
\end{table}

\subsection{N400 and P600 amplitude quantification}
\label{sec:n400_p600_quantification}

ERP components were quantified as trial-level mean amplitudes in predefined post-onset time windows. The N400-window amplitude was measured in the 300--500 ms interval, a standard latency range for semantic processing in visual word comprehension \citep{kutas2011thirty, alday2019much, michaelov2023ignoring}. This window is also consistent with the \texttt{DERCo} validation analysis, which examined predictability-related N400 effects in the 300--500 ms post-stimulus interval \citep{quach2024derco}. Late positive activity was measured in the 500--800 ms interval and is referred to here as the P600-window amplitude. This label denotes a late post-onset positivity window rather than a condition-defined P600 effect.

As described in Section~\ref{sec:data_preprocessing}, the released \texttt{DERCo} epochs were used without additional prestimulus baseline correction. The dependent variables in this analysis are therefore uncorrected mean voltages computed from the released cleaned epochs. The N400 and P600 measures were quantified as mean voltages in predefined post-onset time windows. The N400-window measure was computed as the mean voltage from 300 to 500 ms after word onset, and the P600-window measure was computed as the mean voltage from 500 to 800 ms after word onset. These measures are not baseline-corrected component amplitudes, but window-based voltage summaries derived from the released cleaned epochs.

For each word epoch, participant, and scalp channel, N400-window amplitude was computed as the mean voltage from 300 ms to 499 ms after word onset:

\[
\text{N400}_{\text{amp}} =
\frac{1}{N_{400}}
\sum_{t=300}^{499} V_t ,
\]

\noindent where \(V_t\) is the voltage at time point \(t\) for a given word epoch, participant, and channel. At 1000 Hz, the half-open 300--500 ms interval contains 200 samples, so \(N_{400}=200\).

P600-window amplitude was computed as the mean voltage from 500 ms to 799 ms after word onset:

\[
\text{P600}_{\text{amp}} =
\frac{1}{N_{600}}
\sum_{t=500}^{799} V_t ,
\]

\noindent where \(V_t\) is the voltage at time point \(t\) for a given word epoch, participant, and channel. At 1000 Hz, the half-open 500--800 ms interval contains 300 samples, so \(N_{600}=300\).

These measures are window-based mean amplitudes, not baseline-corrected amplitudes, standardized scores, or condition-level ERP effects. They were used as trial-level dependent variables in the component-level mixed-effects regression analyses. For convenience, we term them N400 size and P600 size in the following sections.

\subsection{Computing word surprisal}

\textit{Word surprisal} is a metric of how unexpected a word is in a given context, calculated as the negative logarithm of the probability of a word given its preceding context:
\[
\boxed{S(w|c) = -\log_2 P(w|c)}
\]
Here $w$ refers to the target word, and $c$ denotes context.

In the present study, the main surprisal predictor was computed with the HuggingFace \texttt{GPT-2} causal language model \citep{radford2019language}. GPT-2 is an autoregressive Transformer model that estimates the probability of each upcoming token from the preceding tokens. This left-to-right architecture makes it suitable for estimating word surprisal during incremental language comprehension, because the model has access only to prior context when assigning a probability to the current token.

GPT-2 operates on byte-pair encoding (BPE) tokens rather than directly on whole words. We therefore first tokenized each sentence using the GPT-2 tokenizer and computed token-level surprisal values in bits using \(-\log_{2} p\), where \(p\) is the probability assigned by the model to the observed token. For words split into multiple BPE tokens, the token-level surprisals were summed to obtain a single word-level surprisal value. Thus, the surprisal of a word corresponds to the summed surprisal of all GPT-2 tokens aligned with that word.

Sentences were processed separately, with a beginning-of-sequence token prepended so that the first word in each sentence also received a model-based surprisal estimate. Punctuation was retained as part of the whitespace-delimited word form. Token-to-word alignment was performed using the tokenizer offset mappings, which allowed each BPE token to be assigned to the corresponding word. The resulting word-level surprisal values were then aligned with the EEG epochs by segment, sentence number, word position, and word form. Surprisal values derived from GPT-Neo were computed only as a robustness check and were not used as the main predictor in the analyses reported in the main text.

\subsection{Computing semantic relevance}

\textit{Contextual semantic relevance} quantifies how strongly a target word is semantically connected to its recent local context. Unlike surprisal, which measures how unexpected a word is given preceding context, contextual semantic relevance estimates the degree of semantic fit between the target word and the words that have just been processed.

The metric was motivated by the assumption that local contextual information is not equally available during incremental comprehension. Recent words tend to exert stronger influence on the processing of the current word than more distant
words \citep{cowan2001magical, oberauer2002access, lewis2005activation, christiansen2016now}. This local-window assumption is also consistent with previous EEG and eye-tracking studies showing that semantic information from a short preceding context can predict neural and reading-time responses
\citep{frank2017word, broderick2018electrophysiological, sun2024attention1}. Considering \texttt{DERCo} using RSVP and RSVP-flanker presentation, contextual semantic relevance was computed only from the three immediately preceding words and the target word, and no following-word or parafoveal-preview term was included.

For each target word \(w_t\), we used pre-trained \texttt{fastText} word embeddings \citep{bojanowski2016enriching} to represent the target word and its three preceding context words, \(w_{t-3}\), \(w_{t-2}\), and \(w_{t-1}\). We first computed the cosine similarity between the target vector and each context-word vector. These target-context similarities estimate how strongly the target word
fits the recent semantic context. To reflect graded contextual availability, the three context words were assigned distance-based weights, with closer words receiving larger weights:

The metric also includes the pairwise semantic similarities among the three preceding context words. These context-context similarities capture the local semantic coherence of the material already processed before the target word is encountered. Each pairwise context-context similarity was assigned a fixed weight of \(0.2\). Thus, the final score combines two sources of information: the
semantic fit between the target word and its preceding context, and the semantic coherence within that preceding context.

Formally, for a target word \(w_t\), semantic relevance is calculated as follows:

\noindent
{\small
\[
\text{Sem\_Rel}(w_t) =
\sum_{k \in \{t-3, t-2, t-1\}}
w_k \cdot \text{Sim}(v_t, v_k)
+
\sum_{(k,l)\in P}
0.2 \cdot \text{Sim}(v_k, v_l)
\]
}

\noindent
where \(v_i\) and \(v_j\) denote the embedding vectors of words \(w_i\) and \(w_j\), \(P=\{(t-3,t-2),(t-3,t-1),(t-2,t-1)\}\), and cosine similarity is defined as:

{\small
\[
\text{Sim}(v_i, v_j) =
\frac{v_i \cdot v_j}{|v_i| |v_j|}
\]
}

\noindent
The context-word weights are defined as:
\[
w_{t-3} = 0.3, \quad
w_{t-2} = 0.6, \quad
w_{t-1} = 0.9.
\]
The pairwise context pairs are \( (t-3, t-2) \), \( (t-3, t-1) \), and \( (t-2, t-1) \), corresponding to the semantic similarities among the three preceding context words.

\begin{figure}[H]
    \vskip -0.06in
    \centering
    \safeincludegraphics[width=0.88\textwidth]{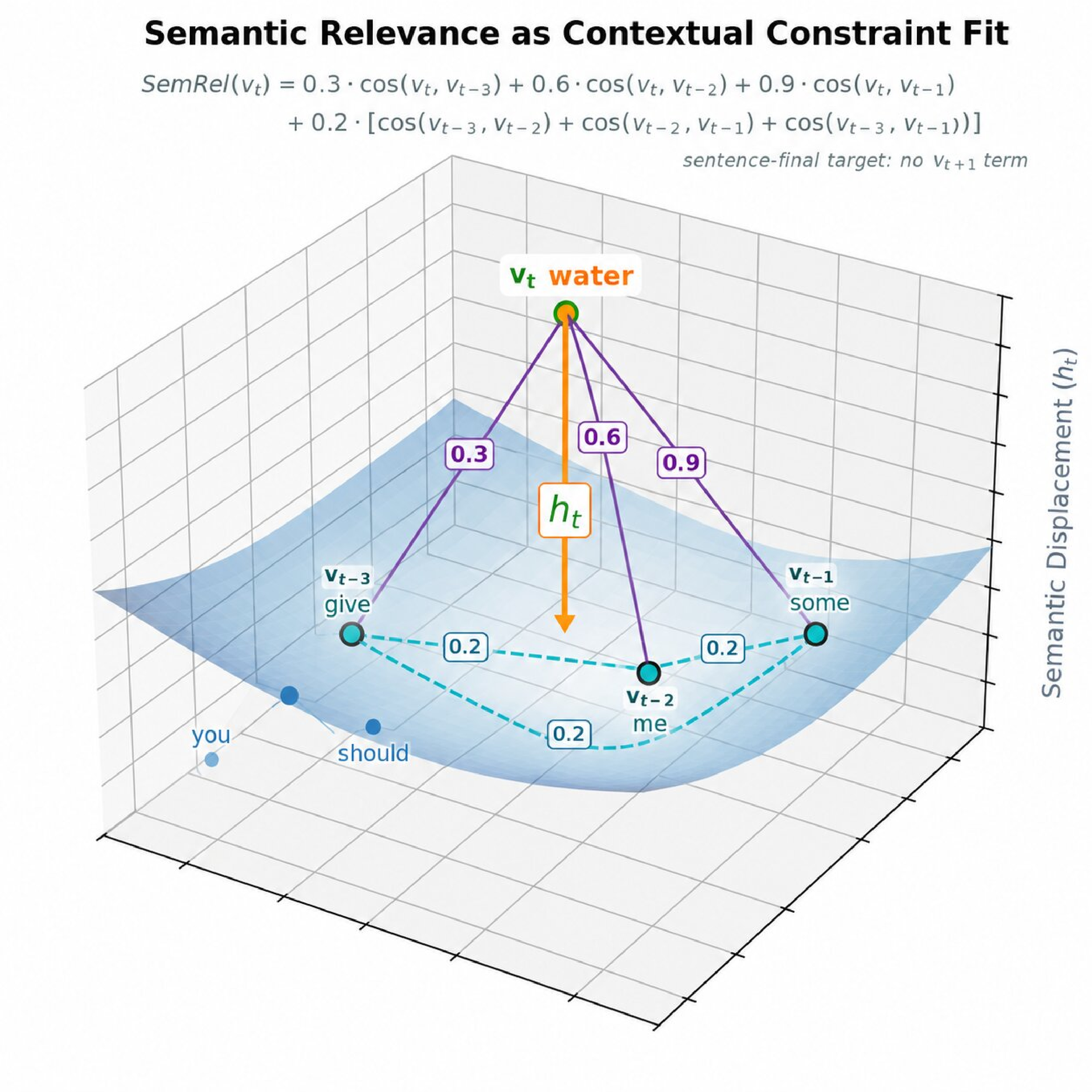}
    \caption{\textbf{Computation of contextual semantic relevance.}
    The figure illustrates the computation of semantic relevance for the target word ``water'' in the sentence ``you should give me some water''. The three preceding words, ``give'' (\(v_{t-3}\)), ``me'' (\(v_{t-2}\)), and `some'' (\(v_{t-1}\)), form the local context. Semantic relevance is computed as a weighted sum of cosine similarities between the target word and these context words, with closer words receiving larger weights (\(0.3\), \(0.6\), and \(0.9\)). Pairwise similarities among the preceding
    context words are weighted by \(0.2\). Because the \texttt{DERCo} stimuli were presented word by word in RSVP, no following-word term was included.}
    \label{fig:semrev}
    \vskip -0.11in
\end{figure}

A higher semantic-relevance value indicates stronger semantic fit between the target word and its recent local context. In the example in Figure~\ref{fig:semrev}, the target word is ``water'', and the context words are ``give'', ``me'', and ``some''. The metric therefore combines the weighted similarities between ``water'' and each preceding word with the pairwise similarities among the preceding words themselves.

The weights used in the metric were fixed \emph{a priori} rather than estimated from the EEG data. The same graded-weighting approach has been used in previous work on eye movements and reading times \citep{sun2023interpretable, sun2024attention1, sun2024attention}. Additional implementation details are provided in Appendix~A.

Semantic relevance and surprisal differ in both computation and interpretation. Surprisal estimates how unexpected a word is given its preceding context. Semantic relevance is relation-based and
integration-oriented, and it estimates how strongly the target word is connected to its recent semantic context. A word can therefore be predictable but only weakly related to the local semantic structure, or relatively unexpected but still semantically well integrated. In the present text stimuli, the weak Pearson's correlation between semantic relevance and surprisal (\(r=-0.10\); see Section~4.1) suggests that the two metrics capture largely distinct aspects of word-level processing, although a low linear correlation alone does not establish full independence. Note that contextual semantic relevance is also simplified as ``semantic relevance' in the present study.

\subsection{MNE rERP regression and GAMM analysis}

To examine how word-level predictors modulate EEG responses, we used two complementary statistical approaches: regression-based ERP analysis (rERP) and generalized additive mixed models (GAMMs). The rERP analysis was used to examine time-resolved predictor effects across channels and time, whereas the GAMM analysis tested predictor effects on mean voltages in predefined N400 and P600 time windows.

First, we conducted the rERP analysis using MNE-Python \citep{Gramfort2014, Smith2015rerp}. This analysis was performed on the released preprocessed word-locked epochs rather than on the continuous EEG signal. For each participant, word epochs were aligned with token-level predictors. Word onset defined time zero for epoching and was not treated as a numerical predictor. At each channel-time sample, the response variable was the single-trial EEG voltage amplitude, denoted as \(EEG(t)\). The design matrix included surprisal, semantic relevance, word length, and log(word frequency). We fitted a full model including both surprisal and semantic relevance, as well as two reduced models in which either semantic relevance or surprisal was excluded.

The rERP coefficients provide time-resolved beta estimates for each predictor, allowing us to examine when and where surprisal and semantic relevance were associated with EEG activity. For semantic relevance and surprisal, channel-by-time statistical maps were corrected using false discovery rate (FDR) correction. Because the analysis was conducted on already epoched data, it should not be interpreted as a full deconvolution model of overlapping responses in the continuous EEG signal. We therefore use the rERP results as complementary time-resolved evidence, while the predefined N400- and P600-window GAMM analyses provide the main inferential tests.

Second, we used GAMMs \citep{wood2017generalized} to test whether the predictors explained variation in component-level ERP responses. N400-window mean voltage (N400 size) was computed from 300--499 ms after word onset, and P600-window mean voltage (P600 size) was computed from 500--699 ms after word onset. Separate GAMMs were fitted for each EEG channel and each ERP window.

GAMMs were used because the relationships between continuous computational predictors and ERP responses may be nonlinear. The models were implemented with the \texttt{mgcv} package in \texttt{R}, which supports smooth terms, random-effect smooths, smoothing-parameter estimation, and AIC-based model comparison \citep{wood2016smoothing, wood2020inference}. Word length and word
frequency were included as lexical control predictors because both are known to influence reading behaviour and ERP responses
\citep{brysbaert2018word,barton2014word,hauk2004effects,hauk2006time,Meulman2023eeg}. Participant and \texttt{token\_id} were included as random-effect smooths to account for repeated observations. Here, \texttt{token\_id} refers to a unique word occurrence at a specific position in the text, rather than to a lexical word type.

For the component-level GAMM analyses, FDR correction was applied using the Benjamini--Hochberg procedure. Correction was performed separately within each predictor \(\times\) ERP-window family across the 32 EEG channels. Thus, separate corrections were applied for semantic relevance in the N400 window, surprisal in the N400 window, semantic relevance in the P600 window, and surprisal in the P600 window. Channel-wise effects were reported as significant when the
FDR-corrected value satisfied \(q < .05\).

To evaluate the incremental contribution of the main predictors, we compared full GAMMs with reduced GAMMs using \(\Delta\)AIC. The Akaike information criterion (AIC) is an estimator of prediction error and thereby relative quality of statistical models for a given set of data. The reduced models excluded either semantic relevance or surprisal while retaining the lexical controls and random-effect structure. These comparisons allowed us to assess whether each main predictor improved model fit beyond the control predictors. We also inspected simpler models without token-level random effects, but these models yielded less conservative inference because repeated observations of the same word token were treated as independent. The models with participant- and token-level random-effect smooths were therefore used as the primary inferential models.

In sum, the rERP and GAMM analyses provide complementary evidence. The rERP analysis estimates time-resolved predictor-related EEG variation across channels and time, whereas the GAMM analysis tests nonlinear effects on N400- and P600- size. 
The main metrics and statistical methods are summarized in Table~\ref{table:sem}.



\begin{table}[H]
	\centering
	\caption{Metrics and statistical analyses}
	\label{table:sem}
	\vspace{0.06in}
	\resizebox{\textwidth}{!}{%
		\begin{tabular}{p{3.2cm} p{3.6cm} p{6.4cm} p{4.2cm}}
			\toprule
			\textbf{Metric} & \textbf{Computation method} & \textbf{Equation} & \textbf{Statistical analysis} \\
			\midrule
			
			Word surprisal &
			Language model, e.g., GPT-2 &
			$\displaystyle -\log_{2} p(w_t \mid w_{<t})$ &
			rERP regression; GAMMs predicting N400 size and P600 size \\
			
			\midrule
			
			Contextual semantic relevance &
			Kernel smoothing + attention-aware &
			$\displaystyle
			\sum_{k \in \{t-3,t-2,t-1\}} 
			\alpha_k\,\mathrm{Sim}(v_t,v_k)
			+
			0.2\sum_{(k,l)\in P}
			\mathrm{Sim}(v_k,v_l)$ &
			rERP regression; GAMMs predicting N400 size and P600 size \\
			
			\bottomrule
		\end{tabular}%
	}
	\vspace{-0.06in}
\end{table}

\section{Results}

\subsection{Results of rERP regression}

Before implementing the statistical analysis, we conducted correlation analysis for the control predictors and main predictors. The correlation analysis showed that surprisal and semantic relevance were only weakly correlated (r = -0.10), indicating that they capture different aspects of the word-level data rather than measuring the same construct. Overall, the correlations among the predictors were generally low to moderate, with no evidence of strong pairwise dependence. This suggests limited collinearity among surprisal, semantic relevance, word frequency, and word length, supporting their simultaneous inclusion in regression analyses as relatively distinct predictors. More is seen in Appendix B.

The rERP regression analysis examined how word-level predictors modulated the time-resolved EEG signal. Across 22 participants and 110 article-level datasets, the analysis included 52,970 word epochs. Of these, 52,918 epochs were successfully aligned with token-level predictor measures, yielding a match rate of 99.90\%.

We fitted three nested rERP regression models. The full model included word onset, word surprisal, semantic relevance, word length, and word frequency. Two reduced models were then fitted by removing one predictor of interest at a time while retaining the remaining lexical covariates. The unique contribution of word surprisal was assessed by comparing the full model with a reduced model excluding surprisal. The unique contribution of semantic relevance was assessed by comparing the full model with a reduced model excluding semantic relevance.

\begin{equation*}
\boxed{
{\scriptsize
\begin{aligned}
\text{full model:}\quad 
EEG(t) &\sim \text{word onset} + \text{surprisal} + \text{semantic relevance} 
+ \text{word length} + \text{log(word frequency)}, \\
\text{reduced model 1:}\quad 
EEG(t) &\sim \text{word onset} + \text{semantic relevance} 
+ \text{word length} + \text{log(word frequency)}, \\
\text{reduced model 2:}\quad 
EEG(t) &\sim \text{word onset} + \text{surprisal} 
+ \text{word length} + \text{log(word frequency)}.
\end{aligned}
}
}
\end{equation*}

Here, \(EEG(t)\) denotes the EEG amplitude at a given time point and channel. The regression coefficients therefore represent time-resolved beta estimates for each predictor across the epoch.

The full rERP model shows that both semantic relevance and surprisal showed reliable rERP effects, but their temporal and spatial profiles differed. Note that we only reported the most significant cases for word surprisal. In the full model, semantic relevance produced its strongest group-level effect in the N400 window, with a peak absolute beta around 389 ms at FT9. Within the 300--500 ms window, the strongest positive semantic-relevance effect was observed at FT9, whereas the strongest negative effect was observed at Pz. The semantic-only model preserved the same general pattern, with strong effects over TP9/FT9 and a corresponding negative effect over Pz. This indicates that the semantic-relevance effect was not dependent on the inclusion of surprisal in the full model.

Surprisal showed a different temporal profile. In the full rERP model, the largest absolute surprisal-related effect occurred later in the epoch, around 750 ms at F7. In the N400 window, the strongest positive surprisal effect was observed at F7, whereas the strongest negative effect was observed at Pz. In the P600 window, surprisal showed stronger frontal positivity, especially at F7, together with a posterior negative effect around Oz. The surprisal-only model showed a similar frontal-positive and posterior-negative pattern.

\begin{figure}[H]
	\centering
	\safeincludegraphics[width=0.85\textwidth]{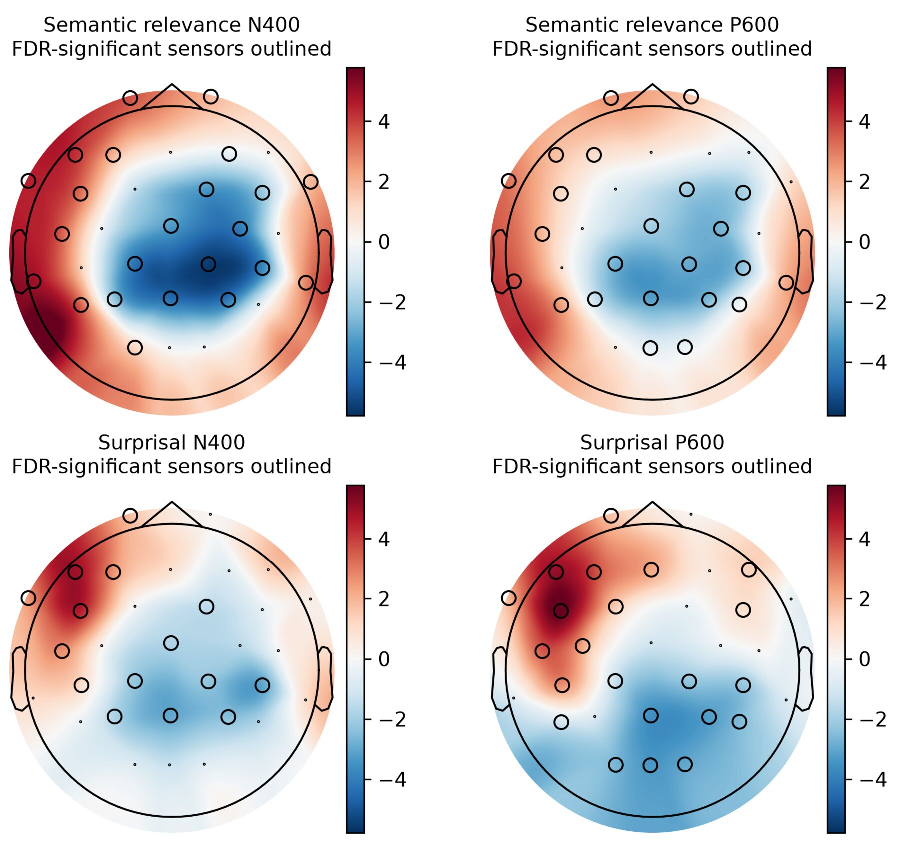}
	\caption{Raw rERP window topographies for semantic relevance and word surprisal. Black circles indicate channels that survived FDR correction.}
	\label{fig:rerp_raw_fdr_topomaps}
\end{figure}

Figure~\ref{fig:rerp_raw_fdr_topomaps} illustrates the spatial distribution of rERP beta effects in the N400 and P600 windows from the full rERP model. The FDR-corrected rERP maps showed time-resolved associations between the predictors and EEG amplitude. These maps are interpreted descriptively and are used to complement the component-level GAMM results. Black circles indicate channels that are significant after FDR correction. Semantic relevance showed broad FDR-significant effects in both windows, particularly in the N400 window. Surprisal showed fewer significant channels in the N400 window, but became more spatially widespread in the P600 window. 

A direct comparison of reduced models suggested that semantic relevance was more strongly associated with N400-window activity, whereas surprisal showed relatively stronger later effects. In the N400 window, the semantic-only model had a larger mean absolute beta across channels than the surprisal-only model. In the P600 window, the surprisal-only model showed a larger mean absolute beta than semantic relevance. 

We next examined the overlap between FDR-significant semantic relevance and surprisal effects in the full model. In the N400 window, 14 channels showed significant effects for both predictors, 9 channels showed a unique semantic-relevance effect, 1 channel showed a unique surprisal effect, and 8 channels showed neither effect. In the P600 window, 16 channels showed effects for both predictors, 7 channels showed a unique semantic-relevance effect, 6 channels showed a unique surprisal effect, and 3 channels showed neither effect. Thus, semantic relevance showed more unique significant channels than surprisal, especially in the N400 window.

For the ROI (region of interest)-level visualization, we used a simplified four scalp region grouping to summarize the dominant anterior--posterior and central--parietal patterns: frontal, central, centro-parietal, and parietal regions. The four scalp regions were defined as follows: frontal = Fp1, Fp2, F3, F4, F7, F8, Fz; central = C3, C4, Cz, T7, T8; centro-parietal = CP1, CP2, CP5, CP6, P3, P4, P7, P8, Pz; parietal/posterior = O1, O2, Oz, TP9, TP10. Information about the 32 EEG channels is provided in Table~\ref{table:eeg_channel_groups} and Figure~\ref{fig:eegmap}.

\begin{figure}[H]
	\centering
	\safeincludegraphics[width=\textwidth]{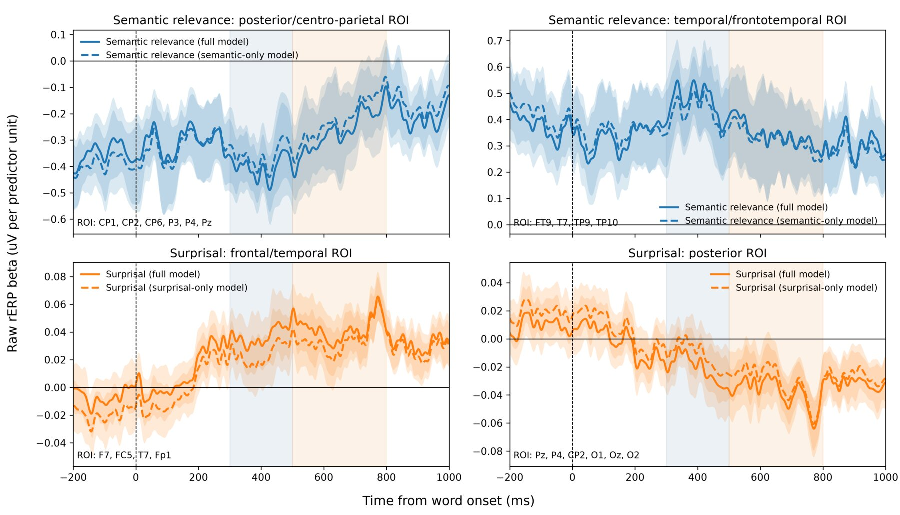}
    \caption{Representative ROI-level rERP beta waveforms with 95\% confidence intervals. Shaded regions mark the N400 and P600 windows. The panels show selected ROI waveforms that illustrate the main temporal patterns for contextual semantic relevance and word surprisal. The full set of ROI-level rERP waveforms across all seven scalp regions is provided in Appendix~C, Figure~\ref{fig:appendix_roi_waveforms}.}
	\label{fig:rerp_roi_waveforms}
\end{figure}

The same pattern was observed at the specific 32 EEG channels. Figure~\ref{fig:rerp_roi_waveforms} shows representative ROI-level rERP beta waveforms with 95\% confidence intervals in the four scalp regions. These selected panels illustrate the main temporal patterns observed for contextual semantic relevance and word surprisal. The complete ROI-level waveform results across all seven scalp regions (grouped in Table~\ref{table:eeg_channel_groups}) are reported in Appendix~C, Figure~\ref{fig:appendix_roi_waveforms}.
Semantic relevance produced sustained effects over posterior, centro-parietal, temporal, and frontotemporal ROIs, with clear modulation in the N400 window. Surprisal showed frontal and temporal positivity together with posterior negativity, and this pattern became more pronounced in the P600 window. 

Figure~\ref{fig:appendix_overlap_dominance} illustrates the effects of all predictors across the N400 and P600 windows (see Appendix C).  Across 7 regions of interest and 2 time windows, semantic relevance was significant in all 14 ROI-window cases in the full model. Surprisal was significant in 13 of 14 ROI-window cases, with the only non-significant case occurring in the occipital ROI during the N400 window. In the reduced models, semantic relevance remained significant in all 14 ROI-window cases, whereas surprisal was significant in 12 of 14 cases. These results indicate that both predictors were reliable, but semantic relevance showed broader spatial robustness. The results are summarized in Table~\ref{table:rerp_fdr_summary}.

\begin{table}[H]
	\centering
	\caption{Summary of FDR-corrected rERP effects for semantic relevance and word surprisal}
	\label{table:rerp_fdr_summary}
	\resizebox{\textwidth}{!}{%
		\begin{tabular}{lllcccc}
			\toprule
			\textbf{Analysis level} & \textbf{Window / Model} & \textbf{Criterion} &
			\textbf{Semantic relevance} & \textbf{Surprisal} &
			\textbf{Both predictors} & \textbf{Neither predictor} \\
			\midrule
			
			Significant channels in full model & N400 & FDR \(q < .05\) &
			23 / 32 & 15 / 32 & -- & -- \\
			
			Significant channels in full model & P600 & FDR \(q < .05\) &
			25 / 32 & 22 / 32 & -- & -- \\
			
			\midrule
			
			Unique and overlapping channels & N400 & FDR \(q < .05\) &
			9 unique & 1 unique & 14 & 8 \\
			
			Unique and overlapping channels & P600 & FDR \(q < .05\) &
			7 unique & 6 unique & 16 & 3 \\
			
			\midrule
			
			Significant ROI-window cases & Full model & FDR \(q < .05\) &
			14 / 14 & 13 / 14 & -- & -- \\
			
			Significant ROI-window cases & Reduced models & FDR \(q < .05\) &
			14 / 14 & 12 / 14 & -- & -- \\
			
			\midrule
			
			Mean absolute beta in reduced models & N400 & Descriptive &
			\(0.113~\mu V\) & \(0.058~\mu V\) & -- & -- \\
			
			Mean absolute beta in reduced models & P600 & Descriptive &
			\(0.079~\mu V\) & \(0.098~\mu V\) & -- & -- \\
			
			\bottomrule
		\end{tabular}%
	}
	\begin{tablenotes}
		\small
		\item Note. FDR effects refer to Benjamini--Hochberg-corrected rERP effects at \(q < .05\). Unique and overlapping channels are reported for FDR-corrected full-model effects. ROI-window cases refer to seven scalp regions across two ERP windows. Mean absolute beta values are descriptive summaries from the reduced rERP models and are reported to compare the relative magnitude of predictor-related beta estimates across channels.
	\end{tablenotes}
\end{table}

In addition to the summary statistics and predictor correlations reported above, Appendix~C provides complementary rERP-based visualizations of the two main predictors. 
Figure~\ref{fig:appendix_sliding_time} summarizes the number of significant channels across time, providing a descriptive view of the temporal extent of each predictor's effect. Figure~\ref{fig:appendix_overlap_dominance} further shows where the two predictors overlap or show predictor-specific effects across the scalp. Finally, Figure~\ref{fig:appendix_model_comparison} compares full and reduced rERP models, and Table~\ref{tab:reduced_beta_summary} reports the corresponding mean absolute beta values. Together, these appendix analyses complement Section~4.1 by showing that contextual semantic relevance and word surprisal are weakly correlated but produce partly distinct temporal, regional, and model-based EEG patterns.

Overall, the rERP results show that semantic relevance and surprisal both modulated EEG responses during naturalistic reading, but with distinguishable profiles. Semantic relevance showed broader and stronger N400-window effects, consistent with contextual semantic integration. Surprisal showed reliable effects in both windows, with comparatively stronger later modulation. These findings complement the GAMM analyses by showing when and where the predictors influenced the continuous EEG signal.

\subsection{Results of GAMMs analysis}


\paragraph{Model specification and rationale.}

The main predictors in the present study, including semantic relevance and surprisal, are item-level variables: each word token has one predictor value and is observed repeatedly across participants. Treating these repeated observations as independent can inflate the effective sample size, reduce standard errors, and produce overly liberal significance tests, a problem related to the language-as-fixed-effect fallacy \citep{clark1973mixed, baayen2008mixed}. Therefore, all primary GAMM analyses included random-effect smooths for both participant and item. The participant random effect accounts for stable between-subject differences in ERP amplitude, whereas the item random effect accounts for repeated observations of the same word token across participants.

We considered a GAMM that included smooth terms for semantic relevance, word surprisal, log (word frequency), and word length, with random-effect smooths for participant and word:

\[
\begin{aligned}
Y \sim\ 
& s(\text{semantic relevance}, k = 5)
+ s(\text{word surprisal}, k = 5) \\
& + s(\log \text{log(word frequency)}, k = 5)
+ s(\text{word length}, k = 5) \\
& + s(\text{participant}, \mathrm{bs}=\text{``re''})
+ s(\text{token\_id}, \mathrm{bs}=\text{``re''}).
\end{aligned}
\]

Here, \texttt{token\_id} refers to a unique word occurrence at a specific position in a specific article, rather than to a lexical word type such as a repeated word form. Because semantic relevance and surprisal are item-level predictors, each word token has one predictor value that is repeated across participants. 
$Y$ denotes either N400 size or P600 size. The notation $s(\cdot)$ denotes a smooth term in a GAMM. In other words, instead of assuming that a predictor has a strictly linear relationship with ERP size, $s(\cdot)$ allows the model to estimate a potentially nonlinear curve from the data.
The parameter $k$ specifies the basis dimension of the smooth. It sets the maximum number of basis functions available to represent the smooth curve. In this study, we used $k = 5$ for the continuous predictors, which allows moderate nonlinearity while avoiding overly flexible curves. The argument $\mathrm{bs}$ specifies the basis type used for a smooth term. For example, $\mathrm{bs}=\text{``re''}$ indicates a random-effect smooth. Thus, $s(\text{token\_id}, \mathrm{bs}=\text{``re''})$ is equivalent to including a random intercept for an unique token, and $s(\text{participant}, \mathrm{bs}=\text{``re''})$ is equivalent to including a random intercept for participant.


\subsubsection{N400 GAMM results}

We first examined whether contextual semantic relevance predicted N400 size beyond surprisaland standard lexical controls. This comparison is central to the present study: surprisal captures probabilistic lexical expectancy, whereas semantic relevance is designed to capture the semantic fit between the current word and the preceding discourse context.

\begin{figure}[H]
    \centering
    \safeincludegraphics[width=\textwidth]{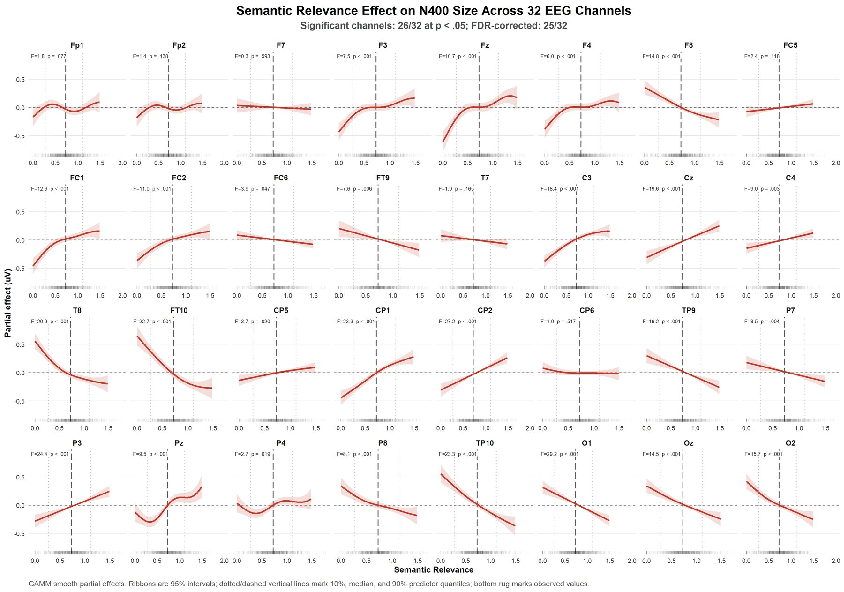}
    \caption{Channel-wise GAMM partial effects of contextual semantic relevance on N400 size across 32 EEG channels. Semantic relevance was significant in 26 of 32 channels at $p < .05$, with 25 channels surviving FDR correction. Note: The \textit{x}-axis represents semantic relevance, and the y-axis represents the model-estimated partial effect on N400-window mean voltage in microvolts (\(\mu V\)), after controlling for word length, log(word frequency), participant variability, and token-level variability.  The red curve shows the fitted smooth effect, and the shaded ribbon indicates the 95\% confidence interval.  A steeper or more curved line indicates a stronger model-estimated association between the predictor and EEG activity, whereas a relatively flat line indicates a weaker association.  The black rug marks along the x-axis show the observed data distribution for the predictor. The vertical dotted lines mark the 10th and 90th percentiles, and the vertical dashed line marks the median of the predictor distribution. The \(F\)-value and associated \(p\)-value in each panel report the approximate significance test for the corresponding GAMM smooth term. These could be applied in reading Figure~\ref{fig:roi_partial_effects}, and Figure~\ref{fig:p600_semrel_all} and figures of partial effects of surprisal in Appendix C.}
    \label{fig:appendix_n400_semrel_all}
\end{figure}

Semantic relevance showed reliable N400 effects across the scalp. In the channel-wise analysis, semantic relevance was significant in 26 of 32 channels at $p < .05$, with 25 channels remaining significant after FDR correction (Figure~\ref{fig:appendix_n400_semrel_all}). These effects were not limited to a narrow electrode cluster, but appeared over frontal, fronto-central, central, centro-parietal, temporal, and posterior channels. Crucially, semantic relevance was evaluated in the same model as surprisal, word frequency, and word length. Significant semantic relevance effects therefore indicate explanatory variance beyond lexical expectancy and basic lexical properties.

Surprisal also showed robust N400 effects. As shown in Figure~\ref{fig:n400_surp_all} in Appendix C, surprisal was significant in 27 of 32 channels, and all 27 channels remaining significant after FDR correction. This confirms the expected relationship between lexical expectancy and N400 activity. However, the semantic relevance results show that probabilistic expectancy is not the whole story: contextual semantic fit remained a reliable predictor even after controlling for surprisal and item-level dependency.

For the ROI-level GAMM visualization, we used a simplified four-region grouping to summarize the dominant anterior--posterior and central--parietal patterns: frontal, central, centro-parietal, and parietal regions, which were defined in Section 4.1. This grouping was used only for the component-level GAMM plots in Figure~\ref{fig:n400_roi_partial}. The broader descriptive scalp grouping in Table~\ref{table:eeg_channel_groups} in Appendix~C, Figure~\ref{fig:appendix_roi_waveforms}, were used for detailed waveform inspection, respectively. The GAMM visualization is consistent with the effects on scalp regions in Figure~\ref{fig:rerp_raw_fdr_topomaps} for rERP analysis. 


The ROI-level analysis provided a descriptive summary of differences in the scalp-level distribution of the two predictors. In these ROI summaries, semantic relevance showed more pronounced estimated effects over central and centro-parietal electrodes, whereas surprisal showed relatively more pronounced effects over frontal and parietal/posterior electrodes.

\begin{figure}[H]
    \centering

    \begin{subfigure}[t]{0.49\textwidth}
        \centering
        \safeincludegraphics[width=\linewidth]{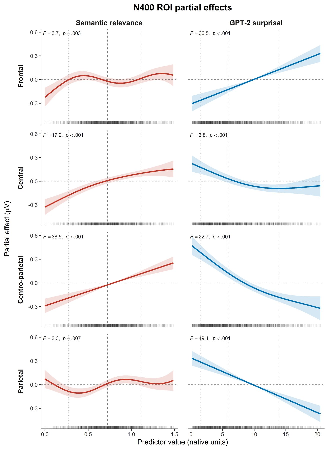}
        \caption{N400 size}
        \label{fig:n400_roi_partial}
    \end{subfigure}
    \hfill
    \begin{subfigure}[t]{0.49\textwidth}
        \centering
        \safeincludegraphics[width=\linewidth]{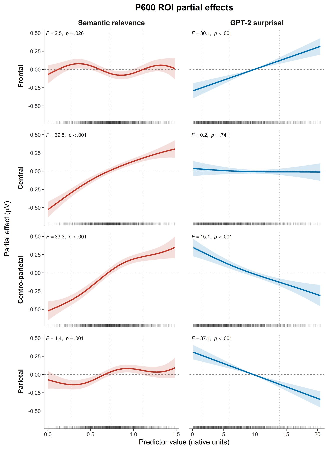}
        \caption{P600 size}
        \label{fig:p600_roi_partial}
    \end{subfigure}


  \caption{ROI-level GAMM partial effects for N400 and P600 size. Panel A shows N400 size, and Panel B shows P600 size. For this visualization, channels were summarized into four broad ROIs: frontal, central, centro-parietal, and parietal/posterior. This simplified grouping was used to display the main spatial pattern of the GAMM effects; the full seven-region rERP waveform grouping is shown in Appendix~C, Figure~\ref{fig:appendix_roi_waveforms}. Red curves show contextual semantic relevance effects, and blue curves show surprisal effects. Models controlled for word length and log word frequency and included random-effect smooths for participant and token\_id. The interpretation of each panel follows the caption of Figure~\ref{fig:appendix_n400_semrel_all}.}
  
    \label{fig:roi_partial_effects}
\end{figure}

In sum, the N400 results support the main hypothesis of this study. surprisal captured the expected contribution of lexical predictability, but semantic relevance provided an additional and statistically reliable account of N400 variation. This indicates that N400 size is shaped not only by how predictable a word is, but also by how semantically relevant it is to the evolving discourse context.

\subsubsection{P600 GAMM results}

We next tested whether semantic relevance and surprisalalso predicted P600 size. This analysis asks whether the contribution of semantic relevance is limited to the N400 window or whether discourse-level semantic fit also influences later post-N400 processing.

\begin{figure}[H]
    \centering
    \safeincludegraphics[width=\textwidth]{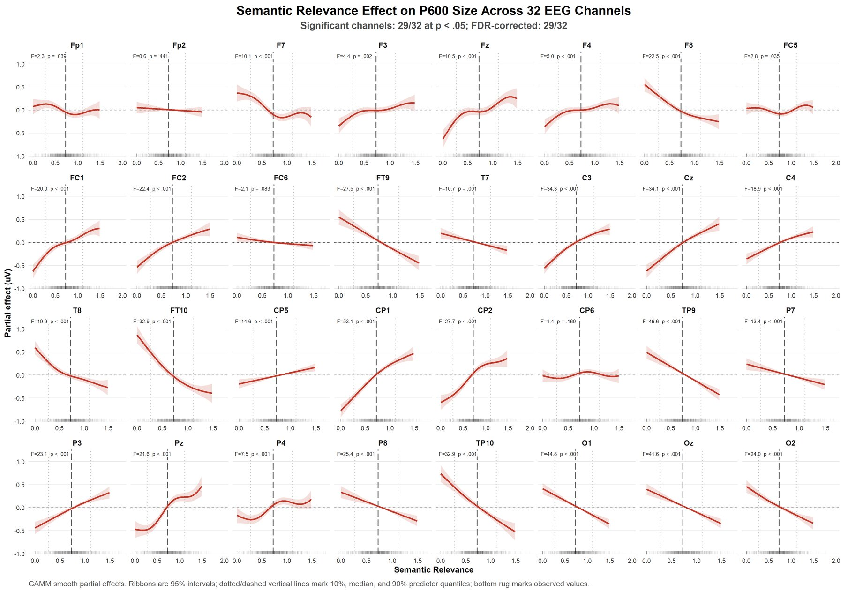}
    \caption{GAMM partial effects of contextual semantic relevance on P600 size across 32 EEG channels. Ribbons indicate 95\% confidence intervals. Vertical dashed and dotted lines mark the median and the 10th/90th percentiles of the predictor distribution; rug marks show observed values.}
    \label{fig:p600_semrel_all}
\end{figure}

Semantic relevance showed especially broad P600 effects. As shown in Figure~\ref{fig:p600_semrel_all}, semantic relevance was significant in 29 of 32 channels at $p < .05$, and all 29 channels remaining significant after FDR correction. This was the most spatially extensive effect among the reported channel-wise analyses. The strongest effects appeared over central, centro-parietal, temporal, and posterior channels. These results suggest that semantic relevance captures not only early semantic fit indexed by N400 size, but also later processes associated with semantic integration, updating, or reanalysis.

Surprisal is also capable of predicting P600 size, but its channel-wise distribution was somewhat less extensive than semantic relevance. Surprisal was significant in 25 of 32 channels, with all 25 channels remaining significant after FDR correction (Appendix C, Figure~\ref{fig:appendix_p600_surp_all}). Thus, both predictors contributed to P600 variation, but semantic relevance showed the broader channel-wise footprint. This contrast is theoretically meaningful: surprisal reflects the unexpectedness of the current word, whereas semantic relevance reflects whether the word is semantically useful or appropriate in the preceding discourse context.

The ROI-level analysis supported this interpretation. As shown in Figure~\ref{fig:p600_roi_partial}, semantic relevance strongly predicted central and centro-parietal activity, with robust effects in the central ROI ($F = 39.5$, $p < .001$) and centro-parietal ROI ($F = 33.3$, $p < .001$). In contrast, surprisal was approximately flat in the central ROI ($F \approx 0$, $p = .942$), but showed strong effects in frontal and parietal regions. This pattern suggests a descriptive difference in scalp-level distribution: semantic relevance tended to show larger estimated effects over central and centro-parietal ROIs, whereas surprisal tended to show larger estimated effects over frontal and parietal/posterior ROIs. 


Overall, the P600 results strengthen the central claim of the study. Semantic relevance was not merely a weak correlate of N400 size, and it showed widespread effects on later ERP activity as well.  Compared with word surprisal, semantic relevance showed a broader P600-window association across channels, with relatively pronounced effects over central and centro-parietal electrodes. This pattern is consistent with the possibility that contextual semantic relevance captures variance related to discourse-level semantic fit beyond lexical prediction.

\paragraph{Summary of semantic relevance and surprisal effects.}

Across both N400- and P600-window amplitudes, semantic relevance and surprisal showed complementary but not identical scalp-level patterns. Semantic relevance tended to show more pronounced effects over central and centro-parietal ROIs, whereas surprisal tended to show relatively stronger effects over frontal and parietal/posterior ROIs. Since these analyses were based on channel-wise and ROI-level scalp models, this pattern should be interpreted as a descriptive difference in scalp distribution rather than as evidence for a stable functional or neural dissociation. The 32-channel smooths are reported as a more fine-grained descriptive check of the ROI-level pattern.
As individual-channel significance counts are less interpretable than the regional organization of effects, the main text focuses on the ROI-level analyses. Overall, the results suggest that semantic relevance and surprisal explain partly different aspects of ERP variation during naturalistic reading. 

Additionally, word length and word frequency also showed reliable effects, although their patterns differed across time windows. In the N400 window, log (word frequency) was significant in 22 of 32 channels, with four additional nominal channels, indicating a broadly distributed frequency effect. Word length showed a weaker but still substantial N400 effect, reaching significance in 17 of 32 channels and nominal significance in six more. In the P600 window, log (word frequency) became even more widespread, with reliable effects in 31 of 32 channels and only Fp2 remaining non-significant. By contrast, word length was more restricted in the P600 range, with reliable effects in 9 of 32 channels and five additional nominal channels. Thus, word frequency showed robust and broadly distributed effects in both windows, whereas word length was strongest in the N400 window and more spatially limited during the P600 interval.

\subsection{GAMM $\Delta$AIC}

We fitted three GAMMs for each analysis: one full model and two nested reduced models. The reduced models were compared against the full model to evaluate the contribution of the predictors of interest. Separate analyses were conducted for N400 size and P600 size across 32 EEG channels. Model improvement was quantified using $\Delta$AIC, defined as
\[
\Delta \mathrm{AIC} = \mathrm{AIC}_{\mathrm{reduced}} - \mathrm{AIC}_{\mathrm{full}}.
\]
Larger positive $\Delta\mathrm{AIC}$ values indicate greater support for the full model over the corresponding reduced model, whereas values closer to or below zero indicate little or no improvement from including the predictor.

\begin{equation*}
\begin{aligned}
\text{Full model:}\quad
Y \sim\ 
& s(\text{semantic relevance}, k = 5) \\
& + s(\text{word surprisal}, k = 5)
+ s(\log \text{log(word frequency)}, k = 5) \\
& + s(\text{word length}, k = 5)
+ s(\text{token\_id}, \mathrm{bs} = \text{``re''}) \\
& + s(\text{participant}, \mathrm{bs} = \text{``re''}).
\end{aligned}
\end{equation*}

\begin{equation*}
\begin{aligned}
\text{Reduced model 1:}\quad
Y \sim\ 
& s(\text{word surprisal}, k = 5)
+ s(\log \text{log(word frequency)}, k = 5) \\
& + s(\text{word length}, k = 5)
+ s(\text{token\_id}, \mathrm{bs} = \text{``re''}) \\
& + s(\text{participant}, \mathrm{bs} = \text{``re''}).
\end{aligned}
\end{equation*}

\noindent
\begin{equation*}
\begin{aligned}
\text{Reduced model 2:}\quad
Y \sim\ 
& s(\text{semantic relevance}, k = 5) \\
 + s(\log \text{log(word frequency)}, k = 5) \\
& + s(\text{word length}, k = 5)
+ s(\text{token\_id}, \mathrm{bs} = \text{``re''}) \\
& + s(\text{participant}, \mathrm{bs} = \text{``re''}).
\end{aligned}
\end{equation*}

The results are summarized in Table~\ref{tab:delta_aic_n400_p600}. Larger positive $\Delta$AIC values indicate greater improvement of the full model over the corresponding reduced model. For N400 size, semantic relevance had the larger relative $\Delta$AIC value in 22 channels, whereas surprisal had the larger relative $\Delta$AIC value in 10 channels. These counts indicate relative dominance between the two predictors, but channels with near-zero or negative $\Delta$AIC values should not be interpreted as providing positive model-improvement evidence. For P600 size, semantic relevance showed stronger overall performance, yielding larger $\Delta$AIC values in 23 channels, compared with 9 channels for surprisal. Detailed channel-level results are provided in Tables~\ref{tab:n400_delta_aic_all_channels} and ~\ref{tab:p600_delta_aic_all_channels}of Appendix~D.

\begin{table}[H]
\centering
\caption{Electrodes with the larger relative $\Delta$AIC value for semantic relevance or surprisal in N400 and P600 size models. Values in parentheses indicate the larger $\Delta$AIC value between the two predictors at the corresponding EEG channel. Positive $\Delta$AIC values indicate that including the predictor improved the full model relative to the reduced model, whereas values near or below zero indicate little or no positive model-improvement evidence.}
\label{tab:delta_aic_n400_p600}
\small

\begin{subtable}{\textwidth}
\centering
\caption{N400 size}
\label{tab:delta_aic_n400}
\begin{tabularx}{\textwidth}{l c X}
\toprule
Stronger predictor & No. of channels & EEG channels with stronger $\Delta$AIC \\
\midrule
Semantic relevance & 21 &
FT10 (58.47), T8 (58.42), CP1 (52.84), TP10 (40.02), Pz (39.14), C3 (36.76), P3 (32.66), CP2 (31.60), F8 (30.54), FC1 (23.73), Fz (23.14), Cz (22.19), O2 (20.85), F4 (17.09), FC2 (15.87), Oz (14.83), O1 (13.25), TP9 (12.99), F3 (12.16), CP5 (4.56), FC6 (1.82) \\
\addlinespace
Surprisal & 10 &
FC5 (48.11), F7 (44.74), T7 (40.17), FT9 (39.13), P4 (38.09), Fp1 (27.41), P8 (19.35), CP6 (17.88), C4 (14.72), Fp2 (10.10) \\
\bottomrule
\end{tabularx}
\end{subtable}

\vspace{1em}

\begin{subtable}{\textwidth}
\centering
\caption{P600 size}
\label{tab:delta_aic_p600}
\begin{tabularx}{\textwidth}{l c X}
\toprule
Stronger predictor & No. of channels & EEG channels with stronger $\Delta$AIC \\
\midrule
Semantic relevance & 23 &
CP1 (101.20), FT10 (86.04), CP2 (85.47), C3 (69.50), Pz (68.92), T8 (64.52), Cz (53.84), FC1 (52.10), TP9 (48.86), F8 (47.82), FT9 (47.77), FC2 (40.90), P3 (39.10), C4 (34.92), P4 (34.52), O1 (30.92), Fz (28.14), CP6 (16.43), Fp1 (15.58), F4 (12.74), CP5 (10.41), TP10 (7.72), P7 (6.94) \\
\addlinespace
Surprisal & 9 &
FC5 (47.92), T7 (38.91), Oz (32.24), F7 (29.95), O2 (29.28), P8 (18.36), F3 (12.88), Fp2 (9.89), FC6 (8.15) \\
\bottomrule
\end{tabularx}
\end{subtable}

\end{table}

Overall, the rERP and GAMM analyses yielded broadly consistent patterns. The rERP visualizations showed scalp-region effects of the main predictors that were in line with the GAMM partial-effect visualizations. Moreover, the GAMM \(\Delta\)AIC analyses provided additional model-comparison evidence for the contribution of surprisal and contextual semantic relevance, further supporting the robustness of the main findings.

\section{Discussion}

\subsection{Summary of main findings}

The present study tested whether word surprisal and semantic relevance explain EEG responses during naturalistic reading, with a focus on N400 and P600 windows. Across both rERP and GAMM analyses, the two predictors showed reliable but partly distinct effects. In the rERP analysis, contextual semantic relevance showed significant effects in 23 of 32 channels for N400 amplitudesand in 25 of 32 channels for P600 amplitudes, whereas word surprisal showed significant effects in 15 channels for N400 and 22 channels for P600. In the GAMM analysis, which allowed nonlinear predictor-response relationships, semantic relevance was significant in 26 channels for N400 size and 29 channels for P600 size, while word surprisal was significant in 25 channels for both components.

These results suggest that neural responses during naturalistic reading are significantly influenced by both lexical prediction and semantic fit. Surprisal captures the conditional probability of a word in context, while semantic relevance captures how well the current word fits the semantic representation built from the preceding context. The convergence between rERP and GAMM results strengthens the central claim that semantic relevance explains ERP variation beyond well-established word surprisal and other predictors.

\subsection{Complementary roles of surprisal and semantic relevance}

The robust effects of surprisal are consistent with a large body of work linking ERP amplitude to semantic expectancy and contextual constraint \citep{michaelov2024strong,Brouwer2021surprisal}. Since the original identification of the N400 as a response to semantic incongruity \citep{kutas1980reading}, many studies have shown that less expected words elicit larger N400 responses \citep{kutas2000electrophysiology,delong2005probabilistic,lau2013dissociating}. Probabilistic accounts of language comprehension therefore predict that word surprisal should explain N400 variation, and our results support this prediction.

However, the present findings also show that word surprisal is not sufficient. Contextual semantic relevance produced broad effects across channels and across both ERP windows. This indicates that the brain is sensitive not only to whether a word is expected, but also to whether it is semantically coherent with the broader context. This distinction is theoretically important because a word can be statistically unexpected but semantically relevant, or statistically likely but weakly related to the developing discourse representation.

The P600 results are especially informative. Although the P600 was first linked to syntactic difficulty \citep{osterhout1992event}, later work has shown that it is also sensitive to semantic anomalies, reanalysis, and discourse-level updating \citep{van2005erp,kuperberg2007neural,brouwer2012getting}. The strong effect of semantic relevance on P600 size supports this broader interpretation. It suggests that P600 activity reflects late integration processes in which the current word is incorporated into an evolving meaning representation, rather than syntactic repair alone.

\subsection{Implications for the distinction between N400 and P600}

Our results support a graded distinction between the N400 and P600. The N400 was more closely associated with early lexical-semantic access and prediction, whereas the P600 was more closely associated with later integration and updating. However, both predictors influenced both components. Thus, N400 and P600 should not be treated as two completely separate stages driven by mutually exclusive mechanisms.

This interpretation is partly consistent with Retrieval-Integration accounts of language processing \citep{brouwer2017neurocomputational,aurnhammer2021retrieval}. Under this view, the N400 primarily reflects retrieval of lexical and semantic information, while the P600 reflects integration of retrieved meaning into a sentence or discourse representation. Our results refine this account by showing that retrieval-related and integration-related signals overlap across time. Prediction and semantic integration both begin early, but their relative contributions change from the N400 to the P600 window.

The rERP and GAMM analyses provide complementary evidence for this claim. The rERP results show that linear effects of surprisal and semantic relevance are distributed across many channels. The GAMM results further show that these effects can be nonlinear, meaning that ERP size does not always change in a simple linear way across the predictor range. This is important for naturalistic language processing, where contextual effects may emerge gradually or only at particular ranges of surprisal and semantic relevance.


The topographic results further support the functional distinction between N400 and P600. N400 effects were generally more central or centro-parietal and negative, consistent with previous descriptions of the N400 distribution \citep{tiedt2020age,aurnhammer2021retrieval}. P600 effects were more posterior and positive, consistent with studies describing the P600 as a late positivity related to integration, updating, and reanalysis \citep{brouwer2017neurocomputational,lopopolo2024tracking,qi2017native}.

Crucially, the effects of word surprisal and contextual semantic relevance were distributed across multiple scalp channels rather than confined to a single electrode. This distributed pattern is expected in naturalistic reading, where prediction, lexical retrieval, semantic integration, and discourse updating interact continuously. Our topographic maps therefore provide spatial support for the main statistical results reported in the rERP and GAMM analyses.

\subsection{Sequential and parallel processing}

The findings contribute to the debate between sequential and parallel accounts of language comprehension. Some recent work argues that lexical retrieval and integration occur sequentially, with the N400 indexing retrieval and the P600 indexing integration \citep{klingvall2024lexical,brouwer2017neurocomputational}. Our results are broadly compatible with this temporal organization, but they also show overlap between prediction and integration effects across both time windows.

This overlap suggests that naturalistic reading involves both sequential and parallel processing. The N400 and P600 have different dominant functions, but neither component is driven by only one type of information. Prediction, lexical retrieval, and semantic integration appear to operate together, with their relative importance changing over time. This interpretation is compatible with predictive processing accounts, in which the brain continuously generates and updates predictions across hierarchical representational levels \citep{clark2013whatever,millidge2021predictive,caucheteux2023evidence}. It is also consistent with the idea that prediction and memory-based integration interact during comprehension \citep{ryskin2023prediction}.




\subsection{Contribution of contextual semantic relevance}

A central contribution of the present study is the proposal and validation of contextual semantic relevance as a crucial EEG predictor. Unlike the well-established surprisal, which measures probabilistic expectancy, contextual semantic relevance measures the semantic fit between the current word and the preceding context. This makes it especially useful for naturalistic reading, where comprehension depends not only on predicting the next word but also on maintaining a coherent discourse representation.

The present EEG results extend previous behavioral and computational work showing that semantic relevance predicts human language processing. Prior studies have shown that surprisal and semantic relatedness metrics can explain reading behavior and processing difficulty \citep{demberg2008data,frank2017word,hale2016information}. More recent work has proposed attention-aware methods for estimating semantic relevance and has shown that these measures predict eye movements during reading \citep{sun2023optimizing,sun2024attention}. Related studies suggest that semantic relevance may also help model speech production, spontaneous speech properties, fMRI responses, and visual attention \citep{sun2026speech,sun2024fmri,sun2024visual}.

The present study adds EEG evidence to this line of work. The broad N400 and P600 effects suggest that semantic relevance is not merely a behavioral predictor, but provides evidence that semantic relevance predicts ERP variation beyond surprisal. 


Despite these, two limitations should be noted. 
First, the present study used channel-wise models rather than a single spatiotemporal hierarchical model. Future work could model the full electrode-time structure directly. 
Second, this study focused on N400 and P600 size. Future work could examine whether contextual semantic relevance also predicts earlier perceptual components, later sustained positivities, or time-resolved EEG dynamics beyond predefined ERP windows.

\section{Conclusion}

This study examined word-aligned EEG responses from the DERCo naturalistic reading dataset, using rERP analysis and GAMMs to test whether surprisal and the proposed semantic relevance metric predict N400 and P600 amplitudes across 32 EEG channels. The findings show that both predictors jointly explain ERP variation during reading, but with partly distinct roles. Surprisal capture prediction-based processing, whereas semantic relevance reveals how well each word fit the evolving discourse representation. The N400 was more closely associated with lexical access and expectancy, while the P600 was more closely associated with later semantic integration and discourse updating. Both predictors influenced both components, suggesting a dynamic language processing system in which prediction and integration operate together across time. By showing that semantic relevance explains ERP variation beyond surprisal, this study validates a computationally interpretable neural predictor and provides new evidence that human language comprehension depends not only on predicting upcoming words, but also on continuously evaluating their semantic fit within context.

\newpage
\appendix

\begin{center}
{\large\bfseries Appendix}
\end{center}

\section*{ Appendix A: Attention-aware approach and its memory capability}
\label{attention}
This section details the rationale behind the attention-aware methodology and its capacity to encapsulate contextual nuances akin to human memory processes. Table~\ref{alg:semantic_relevance} summarizes the main steps of computation.

\begin{table}[t]
\centering
\caption{Computation of contextual semantic relevance}
\label{alg:semantic_relevance}
\small
\begin{tabularx}{\textwidth}{p{0.8cm} X}
\toprule
\textbf{Step} & \textbf{Procedure} \\
\midrule
1 & Represent the target word \(w_t\) and its three preceding context words
\(w_{t-3}\), \(w_{t-2}\), and \(w_{t-1}\) using pre-trained semantic embeddings. \\

2 & Compute cosine similarities between the target vector \(v_t\) and each
context-word vector \(v_{t-3}\), \(v_{t-2}\), and \(v_{t-1}\). \\

3 & Weight the target-context similarities by distance from the target word:
\(\alpha_{t-3}=0.3\), \(\alpha_{t-2}=0.6\), and \(\alpha_{t-1}=0.9\). \\

4 & Compute pairwise cosine similarities among the three preceding context
words and assign each pairwise similarity a weight of \(0.2\). \\

5 & Sum the weighted target-context similarities and the weighted
context-context similarities to obtain the contextual semantic relevance score. \\
\bottomrule
\end{tabularx}
\end{table}

The implementation of the attention-aware methodology unfolds in two principal stages. Initially, semantic parity is ascertained between pairwise terms within a dynamic window. Subsequent to this assessment, a mechanism of weighted calculation is engaged to gauge the relevance of the sentence construct. With the acquisition of these semantic indices, a subsequent layer of weighted values is applied. The aggregation of these weighted semantic values culminates in a composite score, as elaborated in the preceding text. The objective is to accentuate the attention-aware methodology's prowess in contextual encapsulation and to elucidate the intrinsic dynamics of the weighting schema.

As illustrated in Panel A of Fig.~\ref{fig:semrev}, an example is taken to demonstrate the operational dynamics of the ``attention-aware'' approach. Consider the phrase ``I like to eat apple pie'', with ``apple'' as the target word and ``pie'' as the succeeding word. In this instance, the ``attention-aware'' approach deploys a sliding window for contextual analysis, permitting a broader window than conventional methods, thus incorporating a more expansive contextual spectrum. Moreover, the ``attention-aware'' approach is malleable, allowing for adaptation to diverse tasks and facilitating a variegated integration and utilization of contextual data.

\begin{figure*}[!h]
\centering

\begin{subfigure}[t]{0.49\textwidth}
    \centering
    \safeincludegraphics[width=\textwidth]{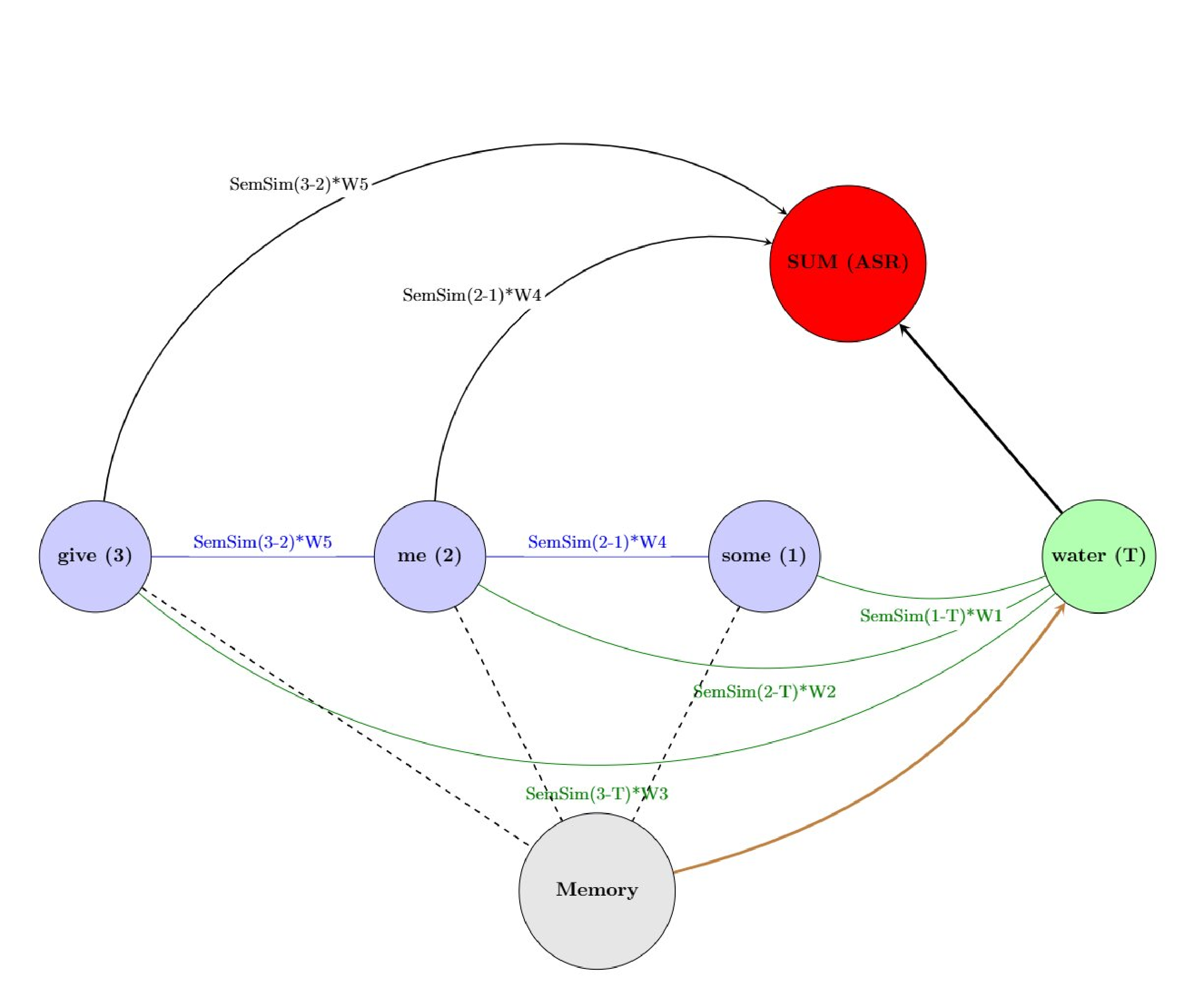}
    \caption{Computation of contextual semantic relevance}
    \label{fig:semrel_comput}
\end{subfigure}
\hfill
\begin{subfigure}[t]{0.49\textwidth}
    \centering
    \safeincludegraphics[width=\textwidth]{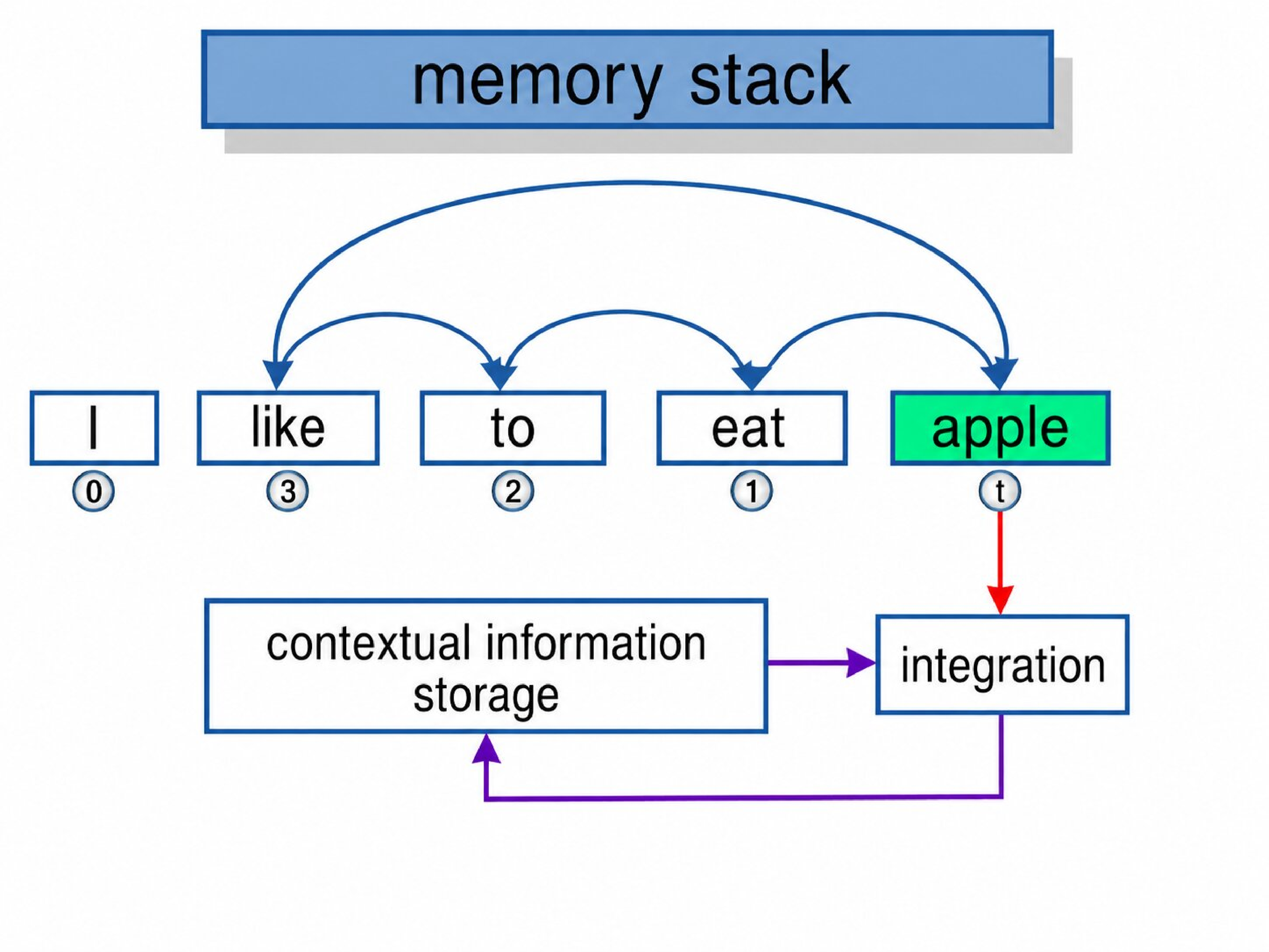}
    \caption{Memory capability and weights in the attention-aware approach}
    \label{fig:memory_capability}
\end{subfigure}

\caption{Computation of semantic relevance and its memory capability and weights adopted in the attention-aware approach. Panel A illustrates the computation of contextual semantic relevance; Panel B shows that the attention-aware method has memory capability.}
\label{fig:memory}

\end{figure*}

As depicted in Fig.~\ref{fig:memory_capability}, within the window stack, terms in proximity to the focal term echo the incipient stages of the forgetting curve, while those further afield reflect the latter phases. To emulate the human forgetting trajectory, a graduated weighting scheme is applied, with greater weights assigned to proximal terms and attenuated weights to distal ones, reflecting the alignment between the human forgetting curve and the attentional weights utilized in this study. The weight values diminish progressively with the spatial chasm between the focal term and the contextual terms (refer to Figure~\ref{fig:memory_capability}), simulate the forgetting curve \citep{loftus1985evaluating}. The attention-aware approach can be correlated with memory models, particularly in how memory decays during information encoding, influencing subsequent linguistic processing during reading. This approach not only actualizes memory functions but also integrates the expectancy effect.

Furthermore, employing a 4-word window (excluding the target word) to simulate the waning of memories over a triad of days, the average rate of memory decline is approximated to one-third daily. This decay is more precipitous initially and decelerates over time. By modeling this decay, we recalibrate the significance of terms based on their remove from the focal term, reducing their value by 0.3. Consequently, terms further from the focal term are accorded lesser weights compared to those in a uniform distribution, establishing a gradual weighting scale, such as 0.3, 0.6, and so on, reflecting their relative proximity to the focal term. Figure~\ref{fig:semrel_comput} provides a two-dimensional illustration of the semantic-relevance computation, complementing the three-dimensional example shown in Figure~\ref{fig:semrev}. 

Panel B of Fig.~\ref{fig:memory} illustrates how our adopted attention-aware approach simulates a short-term memory stack, reflecting the retention of prior encountered terms and their significations. The method of applying varied weights of semantic relevance between words share similarity with the attention mechanisms observed in Transformers. The attention-aware approach not only facilitates the effective assimilation of contextual information but also enables memory storage, retrieval, and integration.

In conclusion, as previously noted, the ``attention-aware'' method resonates with the attention algorithms in Transformers, primarily due to the computational efficiency enhancements achieved through contextual information incorporation. However, our approach does not incorporate any attention layers from Transformers. Each computational step in the attention-aware metrics is transparent and interpretable.

\section*{Appendix B: Correlations among various predictors}

\begin{figure}[H]
    \centering
    \safeincludegraphics[width=0.8\textwidth]{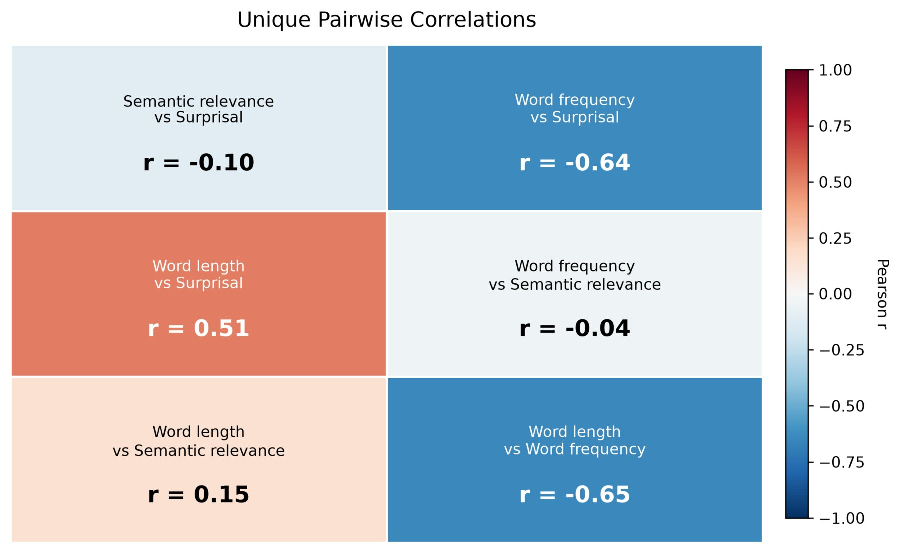}
    \caption{Correlation heatmap for the predictors in EEG data.}
    \label{fig:cor}
\end{figure}

Figure~\ref{fig:cor} presents the pairwise Pearson correlations among the word-level predictors used in the regression analyses: surprisal, semantic relevance, word frequency, and word length. The correlation between surprisal and semantic relevance was very small (r = -0.10), suggesting that these two variables reflect largely distinct dimensions of lexical processing rather than overlapping measures. Other correlations were also within an acceptable range: surprisal was negatively correlated with word frequency (r = -0.64) and positively correlated with word length (r = 0.51), while word frequency and word length were negatively correlated (r = -0.65), consistent with longer words tending to be less frequent. Importantly, no pairwise correlation approached a level that would indicate severe multicollinearity, suggesting that the predictors can be included together in regression models with relatively low risk of collinearity-related instability.

\section*{Appendix C: Supplementary Results}

This section provides additional information about the temporal profile, spatial distribution, and model robustness of the rERP effects reported in the main text. These analyses are included to show that the effects of contextual semantic relevance and word surprisal were not confined to isolated channels or single time points, but were distributed across time and scalp regions.

Table~\ref{tab:appendix_models} summarizes the two statistical analysis methods used in the present study. It compares their response variables, control predictors, and other key components. Overall, these two methods provide complementary evidence and allow the results to be cross-validated, thereby strengthening the validity and robustness of the findings.

\begin{table}[H]
\centering
\caption{Statistical models used in the rERP and GAMM analyses.}
\label{tab:appendix_models}
\small
\begin{tabularx}{\textwidth}{l X}
\toprule
Analysis & Model specification \\
\midrule
rERP full model & EEG amplitudes $\sim$ word surprisal + semantic relevance + word length + log word frequency \\
rERP reduced models & Separate models excluding either word surprisal or semantic relevance \\
GAMM model & \makecell{ERP size \\(mean ERP amplitude)} $\sim s$(word length) + $s$(log word frequency) + $s$(word surprisal) + $s$(semantic relevance) + $s$(word, bs = ``re'') + $s$(participant, bs = ``re'') \\
\bottomrule
\end{tabularx}
\end{table}

\begin{figure}[H]
    \centering
    \safeincludegraphics[width=\textwidth]{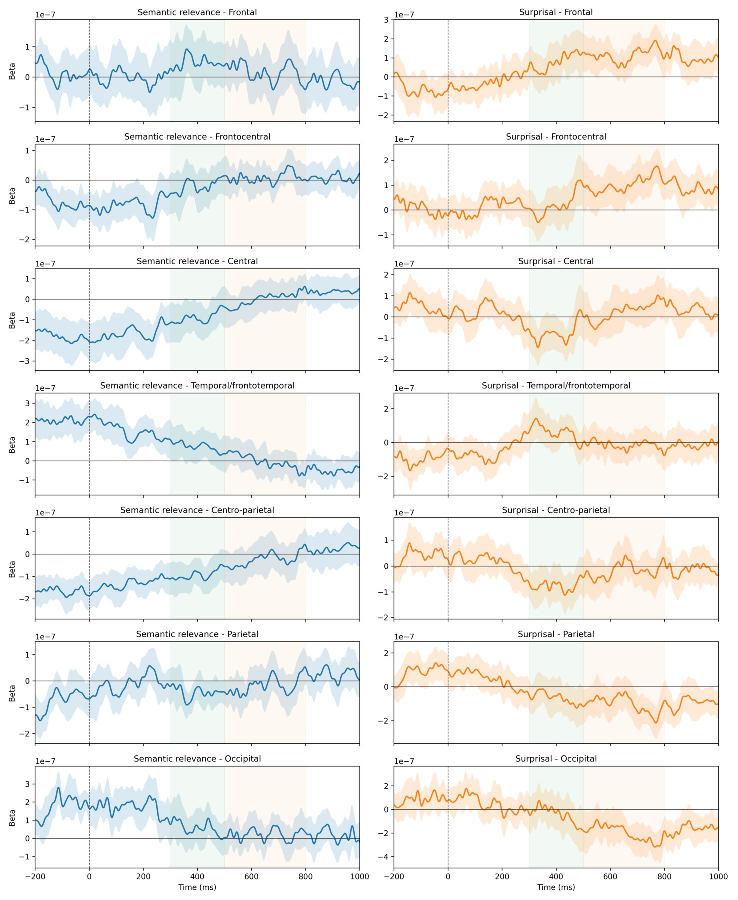}
    \caption{ROI-level rERP waveforms with 95\% confidence intervals for contextual semantic relevance and word surprisal. The waveforms show the temporal evolution of predictor effects across frontal, fronto-central, central, temporal, centro-parietal, parietal, and occipital regions. These plots support the main analysis by showing that semantic relevance and word surprisal produced temporally extended effects rather than isolated point estimates.}
    \label{fig:appendix_roi_waveforms}
\end{figure}

Figure~\ref{fig:appendix_roi_waveforms} illustrates the regional time course of rERP effects. Contextual semantic relevance and word surprisal both showed effects across multiple ROIs, but their temporal and regional profiles differed. These waveform-level results complement the channel-wise analyses reported in the main text and provide visual evidence for distributed ERP effects during naturalistic reading.

\begin{figure}[H]
    \centering
    \safeincludegraphics[width=\textwidth]{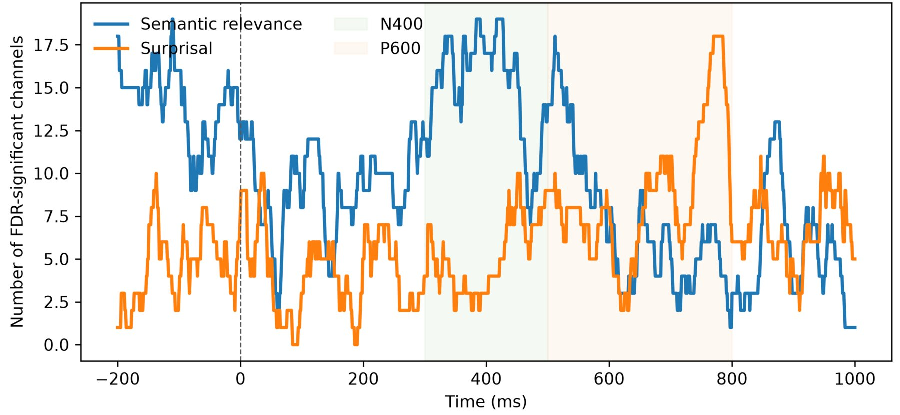}
    \caption{Sliding-time analysis showing the number of significant EEG channels for contextual semantic relevance and word surprisal across time. The shaded windows indicate the N400 and P600 analysis intervals. Solid lines show the full-model effects, and dashed lines show the corresponding reduced-model effects.}
    \label{fig:appendix_sliding_time}
\end{figure}

Figure~\ref{fig:appendix_sliding_time} illustrates how broadly each predictor affected EEG activity over time. Contextual semantic relevance showed a high number of significant channels in the N400 window and remained influential into the later time range. Word surprisal showed a more variable temporal profile, with effects appearing in both early and late windows. This analysis supports the interpretation that prediction and semantic integration both contribute to ERP dynamics across time.

\begin{figure}[H]
    \centering
    \safeincludegraphics[width=\textwidth]{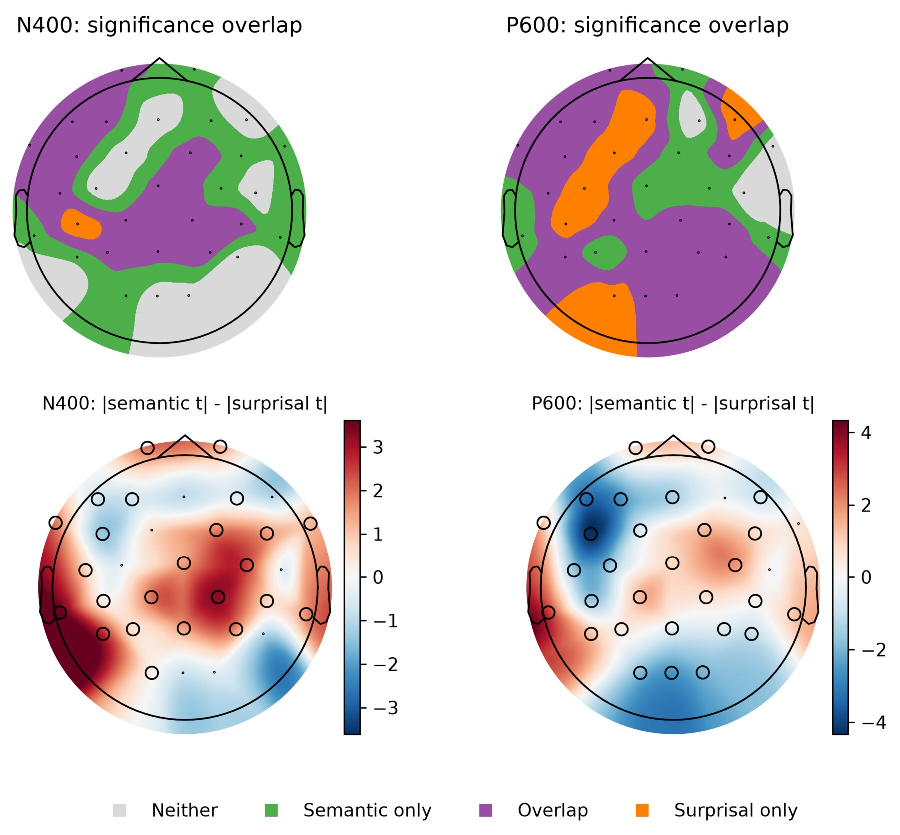}
    \caption{Overlap and dominance maps for contextual semantic relevance and word surprisal in the N400 and P600 windows. The upper panels show channels where only semantic relevance, only word surprisal, both predictors, or neither predictor reached significance. The lower panels show the relative dominance of the two predictors, computed as the absolute semantic relevance effect minus the absolute word surprisal effect.}
    \label{fig:appendix_overlap_dominance}
\end{figure}

Figure~\ref{fig:appendix_overlap_dominance} illustrates the spatial relationship between the two main predictors. The overlap maps show that semantic relevance and word surprisal shared effects in some channels but also had predictor-specific regions. The dominance maps further show that their relative effect strengths differed across scalp locations and ERP windows. This pattern supports the claim that semantic relevance and word surprisal are complementary rather than redundant predictors.

\begin{figure}[H]
    \centering
    \safeincludegraphics[width=\textwidth]{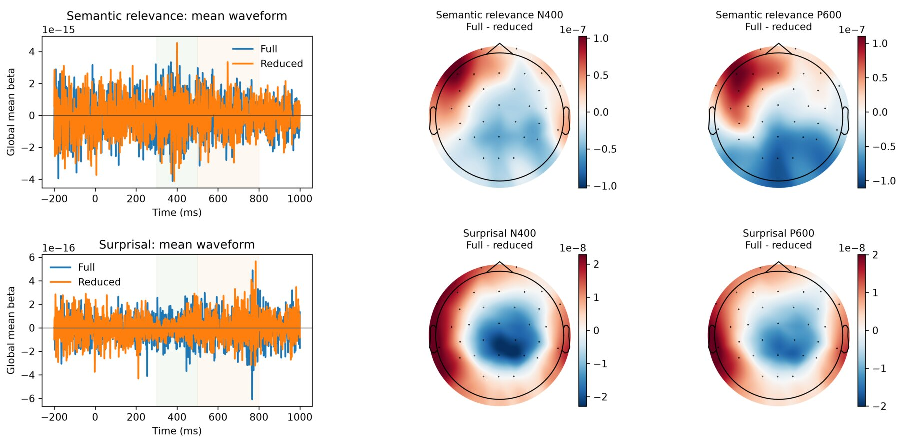}

 \caption{Comparison between full and reduced rERP models. The left panels show grand mean fitted beta waveforms from the full and reduced models. The right panels show topographic full-minus-reduced differences in the N400 and P600 windows. Because the plotted quantities are model-estimate differences rather than raw ERP amplitudes, the numerical scale should be interpreted as relative model-estimate change. Descriptive mean absolute beta values are reported separately in the text or accompanying metrics Tabl~\ref{tab:reduced_beta_summary}.}
    \label{fig:appendix_model_comparison}
\end{figure}

Figure~\ref{fig:appendix_model_comparison} provides additional model-comparison evidence. The full and reduced models produced broadly similar waveforms, but their topographic differences indicate that each predictor contributed unique information to the estimated ERP effects. This supports the main conclusion that contextual semantic relevance explains ERP variation beyond word surprisal, and that the two predictors should be modeled jointly.

\begin{table}[H]
\centering
\caption{Mean absolute rERP beta values across channels in reduced models.}
\label{tab:reduced_beta_summary}
\begin{tabular}{lll}
\toprule
Window & Predictor/model & Mean absolute beta \\
\midrule
N400 & Semantic relevance-only model & 0.113 $\mu$V \\
N400 & Surprisal-only model & 0.058 $\mu$V \\
P600 & Semantic relevance-only model & 0.079 $\mu$V \\
P600 & Surprisal-only model & 0.098 $\mu$V \\
\bottomrule
\end{tabular}
\end{table}



\begin{figure}[H]
    \centering
    \safeincludegraphics[width=\textwidth]{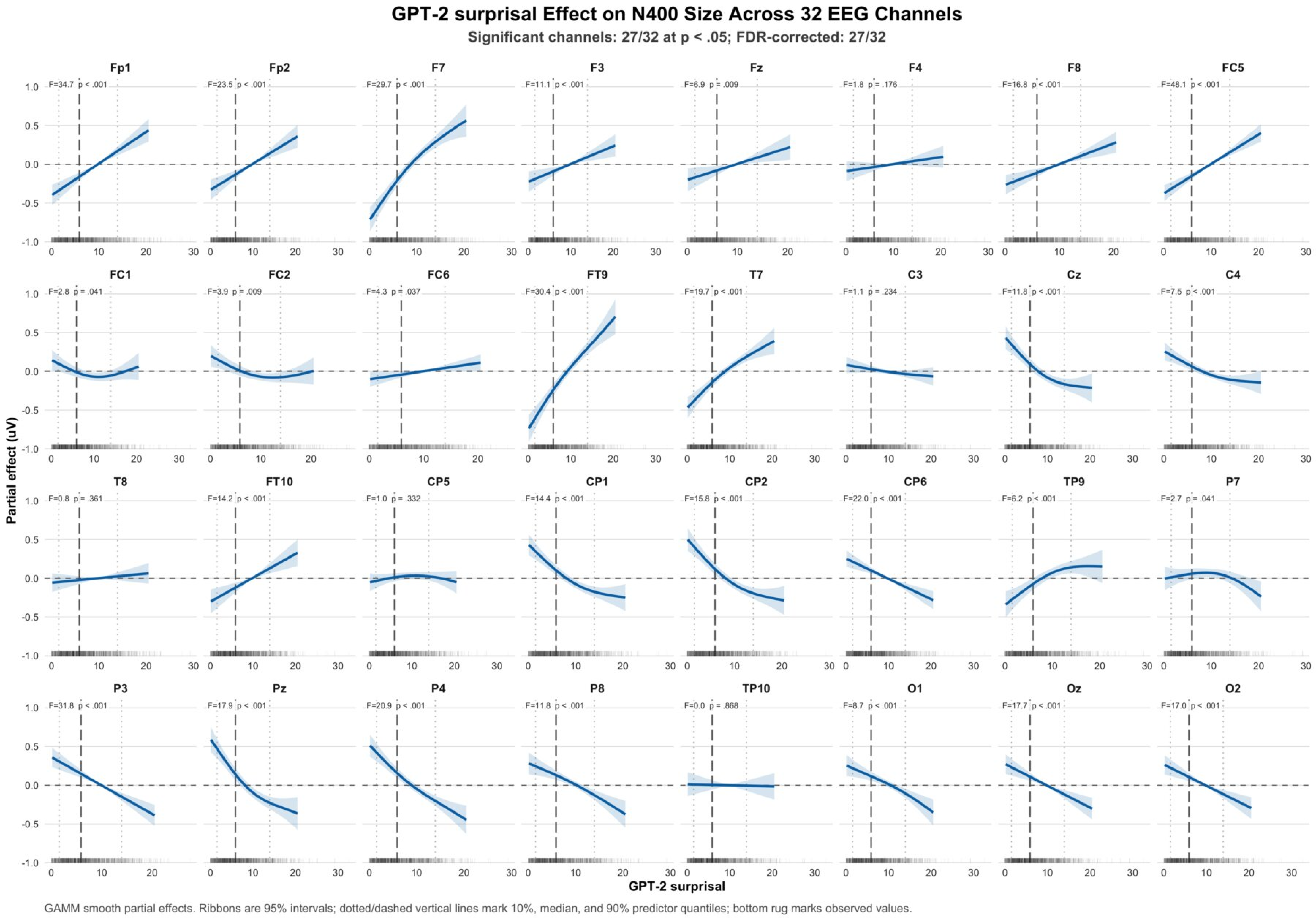}
    \caption{GAMM partial effects of surprisal on N400 size across 32 EEG channels. Ribbons indicate 95\% confidence intervals. Vertical dashed and dotted lines mark the median and the 10th/90th percentiles of the predictor distribution; rug marks show observed values.}
    \label{fig:n400_surp_all}
\end{figure}

\begin{figure}[H]
    \centering
   \safeincludegraphics[width=\textwidth]{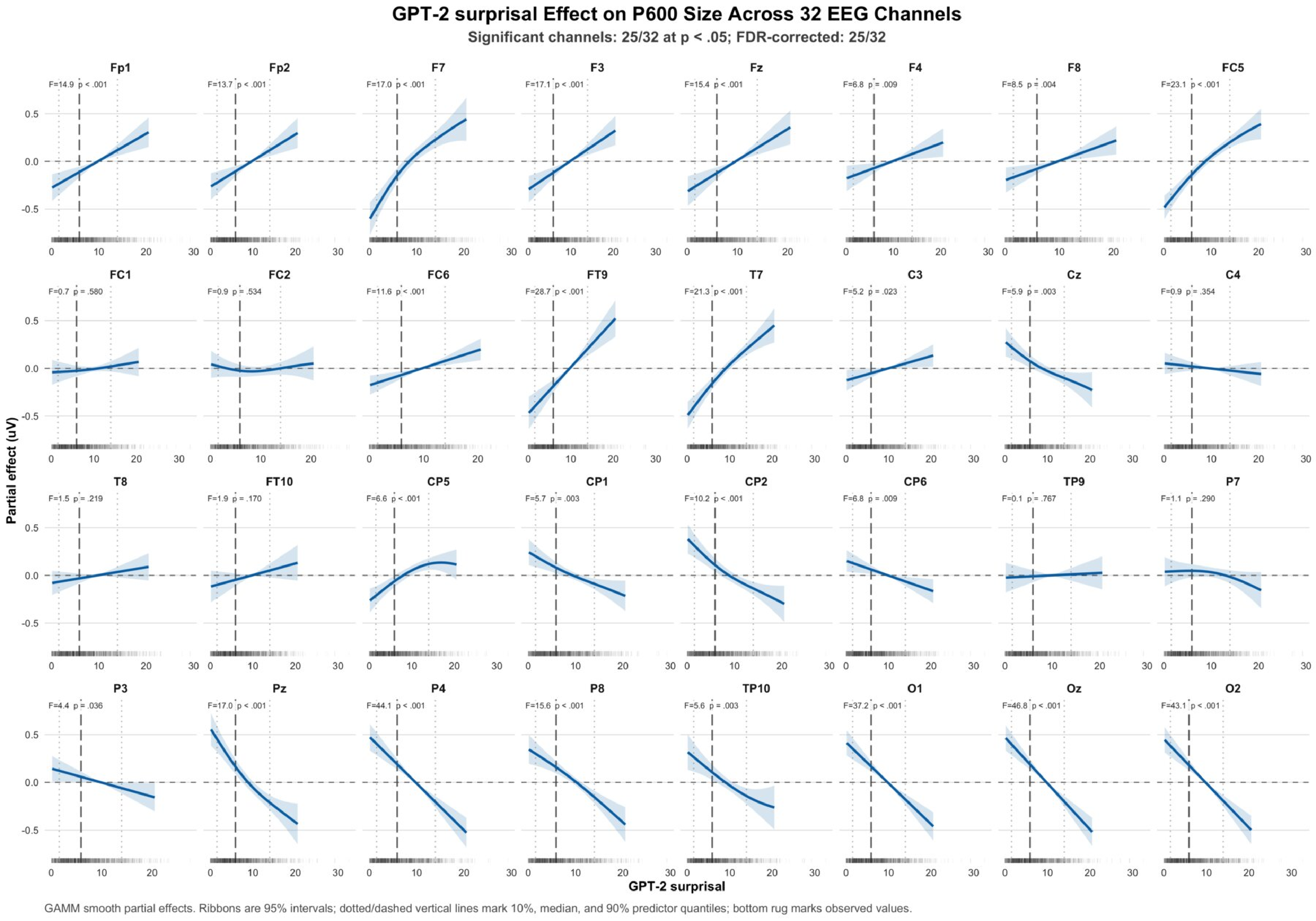}
    \caption{Additional channel-wise GAMM partial effects of surprisal on P600 size across 32 EEG channels. Surprisal was significant in 25 of 32 channels at $p < .05$, with all 25 channels surviving FDR correction.}
   \label{fig:appendix_p600_surp_all}
\end{figure}

Surprisal also showed robust N400 effects. As shown in Figure~\ref{fig:n400_surp_all}, surprisal was significant in 27 of 32 channels, and all 25 channels remaining significant after FDR correction. This confirms the expected relationship between lexical expectancy and N400 activity. However, the semantic relevance results show that probabilistic expectancy is not the whole story: contextual semantic fit remained a reliable predictor even after controlling for surprisal and item-level dependency.  However, as for P600 size, surprisal was significant in 25 of 32 channels at $p < .05$, with all 25 channels surviving FDR correction, which is shown in Figure~\ref{fig:appendix_p600_surp_all}.

\section*{Appendix D: $\Delta$AIC of GAMMs}

To further compare the relative contribution of surprisal and contextual semantic relevance, we computed $\Delta$AIC values by removing each predictor from the full channel-wise GAMM. In the N400 size models, semantic relevance showed larger \(\Delta\)AIC values in many channels, with relatively larger values often appearing over central and centro-parietal electrodes. Surprisal showed larger \(\Delta\)AIC values in a smaller set of frontal, fronto-central, and parietal/posterior electrodes
(Table~\ref{tab:n400_delta_aic_all_channels}). A similar but more pronounced pattern was observed in the P600 size models, where semantic relevance again provided stronger model improvement across most channels, particularly over central and centro-parietal regions, while surprisal was stronger only in selected frontal, fronto-central, and posterior channels (Table~\ref{tab:p600_delta_aic_all_channels}). Overall, these results suggest that both predictors contribute to ERP component size, but semantic relevance provides broader and more consistent explanatory support across channels.

\begin{table}[htbp]
\centering
\caption{$\Delta$AIC values for surprisal and semantic relevance across EEG channels in the N400 size models. Bold values indicate the larger relative $\Delta$AIC within each channel, not necessarily positive model support. Positive $\Delta$AIC values indicate stronger improvement of the full model when the predictor is included; values near or below zero indicate little or no improvement.}
\label{tab:n400_delta_aic_all_channels}
\scriptsize
\begin{threeparttable}
\begin{tabular}{lrrl}
\toprule
Channel & $\Delta$AIC surprisal & $\Delta$AIC SemRel & Larger relative predictor \\
\midrule
C3 & -2.56 & \textbf{36.76} & SemRel \\
C4 & \textbf{14.72} & 8.68 & Surprisal \\
CP1 & 13.27 & \textbf{52.84} & SemRel \\
CP2 & 27.08 & \textbf{31.60} & SemRel \\
CP5 & -2.40 & \textbf{4.56} & SemRel \\
CP6 & \textbf{17.88} & -4.99 & Surprisal \\
Cz & 16.82 & \textbf{22.19} & SemRel \\
F3 & 8.82 & \textbf{12.16} & SemRel \\
F4 & -1.03 & \textbf{17.09} & SemRel \\
F7 & \textbf{44.74} & -6.36 & Surprisal \\
F8 & 9.08 & \textbf{30.54} & SemRel \\
FC1 & -1.90 & \textbf{23.73} & SemRel \\
FC2 & 1.41 & \textbf{15.87} & SemRel \\
FC5 & \textbf{48.11} & -2.16 & Surprisal \\
FC6 & 1.49 & \textbf{1.82} & SemRel \\
FT10 & 8.51 & \textbf{58.47} & SemRel \\
FT9 & \textbf{39.13} & 7.43 & Surprisal \\
Fp1 & \textbf{27.41} & 5.53 & Surprisal \\
Fp2 & \textbf{10.10} & -2.43 & Surprisal \\
Fz & 2.71 & \textbf{23.14} & SemRel \\
O1 & 11.96 & \textbf{13.25} & SemRel \\
O2 & 17.71 & \textbf{20.85} & SemRel \\
Oz & 10.57 & \textbf{14.83} & SemRel \\
P3 & 13.76 & \textbf{32.66} & SemRel \\
P4 & \textbf{38.09} & 13.43 & Surprisal \\
P7 & -4.13 & \textbf{-0.53} & SemRel \\
P8 & \textbf{19.35} & 10.94 & Surprisal \\
Pz & 36.65 & \textbf{39.14} & SemRel \\
T7 & \textbf{40.17} & 1.08 & Surprisal \\
T8 & -1.49 & \textbf{58.42} & SemRel \\
TP10 & -1.66 & \textbf{40.02} & SemRel \\
TP9 & 5.94 & \textbf{12.99} & SemRel \\
\bottomrule
\end{tabular}

\begin{tablenotes}
\footnotesize
\item Note. $\Delta$AIC was computed as $\mathrm{AIC}_{\mathrm{reduced}} - \mathrm{AIC}_{\mathrm{full}}$. Larger positive values indicate stronger improvement of the full model when the predictor is included. SemRel = semantic relevance.
\end{tablenotes}
\end{threeparttable}
\end{table}

\begin{table}[htbp]
\centering
\caption{$\Delta$AIC values for surprisal and semantic relevance across EEG channels in the P600 size models. Bold values indicate the larger $\Delta$AIC within each channel. SemRel = semantic relevance}
\label{tab:p600_delta_aic_all_channels}
\scriptsize
\begin{threeparttable}
\begin{tabular}{lrrl}
\toprule
Channel & $\Delta$AIC surprisal & $\Delta$AIC SemRel & Stronger predictor \\
\midrule
C3 & 3.73 & \textbf{69.50} & SemRel \\
C4 & -1.90 & \textbf{34.92} & SemRel \\
CP1 & 3.34 & \textbf{101.20} & SemRel \\
CP2 & 12.27 & \textbf{85.47} & SemRel \\
CP5 & 9.46 & \textbf{10.41} & SemRel \\
CP6 & 5.09 & \textbf{16.43} & SemRel \\
Cz & 1.87 & \textbf{53.84} & SemRel \\
F3 & \textbf{12.88} & 8.76 & Surprisal \\
F4 & 4.66 & \textbf{12.74} & SemRel \\
F7 & \textbf{29.95} & 29.24 & Surprisal \\
F8 & 4.91 & \textbf{47.82} & SemRel \\
FC1 & -0.94 & \textbf{52.10} & SemRel \\
FC2 & -4.22 & \textbf{40.90} & SemRel \\
FC5 & \textbf{47.92} & 2.83 & Surprisal \\
FC6 & \textbf{8.15} & 1.24 & Surprisal \\
FT10 & -0.59 & \textbf{86.04} & SemRel \\
FT9 & 12.03 & \textbf{47.77} & SemRel \\
Fp1 & 12.81 & \textbf{15.58} & SemRel \\
Fp2 & \textbf{9.89} & 7.33 & Surprisal \\
Fz & 10.18 & \textbf{28.14} & SemRel \\
O1 & 27.92 & \textbf{30.92} & SemRel \\
O2 & \textbf{29.28} & 27.26 & Surprisal \\
Oz & \textbf{32.24} & 29.70 & Surprisal \\
P3 & 1.93 & \textbf{39.10} & SemRel \\
P4 & 32.35 & \textbf{34.52} & SemRel \\
P7 & -4.52 & \textbf{6.94} & SemRel \\
P8 & \textbf{18.36} & 16.42 & Surprisal \\
Pz & 31.36 & \textbf{68.92} & SemRel \\
T7 & \textbf{38.91} & 7.93 & Surprisal \\
T8 & -0.49 & \textbf{64.52} & SemRel \\
TP10 & -36.31 & \textbf{7.72} & SemRel \\
TP9 & -1.75 & \textbf{48.86} & SemRel \\
\bottomrule
\end{tabular}

\begin{tablenotes}
\footnotesize
\item Note. $\Delta$AIC was computed as $\mathrm{AIC}_{\mathrm{reduced}} - \mathrm{AIC}_{\mathrm{full}}$. Larger positive values indicate stronger improvement of the full model when the predictor is included. SemRel = semantic relevance.
\end{tablenotes}
\end{threeparttable}
\end{table}

\end{document}